\documentclass[10pt,twocolumn,letterpaper]{article}

\usepackage{algorithm}
\usepackage{algpseudocode}
\usepackage{multirow}
\usepackage{multicol}
\usepackage{colortbl}
\usepackage[utf8]{inputenc} % allow utf-8 input
\usepackage[T1]{fontenc}    % use 8-bit T1 fonts
\usepackage{booktabs}       % professional-quality tables
\usepackage{amsmath}        % gather, align, etc.
\usepackage{amsfonts}       % blackboard math symbols
\usepackage{nicefrac}       % compact symbols for 1/2, etc.
\usepackage{microtype}      % microtypography
\usepackage{xcolor}         % colors
\usepackage{wrapfig} 
\usepackage{makecell}
\usepackage{titletoc}
\usepackage{wrapfig}
\usepackage{booktabs}
\usepackage{algorithm}
\usepackage{algpseudocode}
\algrenewcommand\algorithmicrequire{\textbf{Input:}}
\algrenewcommand\algorithmicensure{\textbf{Output:}}
\usepackage{marvosym}
\usepackage{lipsum}

\usepackage[svgnames]{xcolor}

\usepackage[pagenumbers]{cvpr} % To force page numbers, e.g. for an arXiv version

\definecolor{cvprblue}{rgb}{0.21,0.49,0.74}
\usepackage[pagebackref,breaklinks,colorlinks,allcolors=cvprblue]{hyperref}

\def\paperID{} % *** Enter the Paper ID here
\def\confName{CVPR}
\def\confYear{2026}

\title{EgoDemoGen: Egocentric Demonstration Generation for Viewpoint Generalization in Robotic Manipulation}

\author{
    Yuan Xu\textsuperscript{\rm 1,2}\footnotemark[1]\quad
    Jiabing Yang\textsuperscript{\rm 1,2}\footnotemark[1]\quad
    Xiaofeng Wang\textsuperscript{\rm 3,4}\quad
    Yixiang Chen\textsuperscript{\rm 1,2}\quad \\
    Zheng Zhu\textsuperscript{\rm 3}\footnotemark[2]\quad 
    ~Bowen Fang\textbf{\textsuperscript{\rm 1,2}}\quad
    Guan Huang\textbf{\textsuperscript{\rm 3}}\quad
    Xinze Chen\textbf{\textsuperscript{\rm 3}}\quad
    Yun Ye\textbf{\textsuperscript{\rm 3}}\quad \\
    Qiang Zhang\textbf{\textsuperscript{\rm 5}}\quad 
    ~Peiyan Li\textbf{\textsuperscript{\rm 1,2}}\quad
    Xiangnan Wu\textbf{\textsuperscript{\rm 1,2}}\quad
    Kai Wang\textbf{\textsuperscript{\rm 2}}\quad
    Bing Zhan\textbf{\textsuperscript{\rm 1,2}}\quad \\
    Shuo Lu\textbf{\textsuperscript{\rm 1,2}}\quad 
    ~Jing Liu\textbf{\textsuperscript{\rm 1,2,6}}\quad
    Nianfeng Liu\textbf{\textsuperscript{\rm 1,2,6}}\quad
    Yan Huang\textbf{\textsuperscript{\rm 1,2,6}}\footnotemark[2]\quad
    Liang Wang\textbf{\textsuperscript{\rm 1,2}}\footnotemark[2]\quad \\
    ~\textsuperscript{\rm 1}UCAS 
    ~\textsuperscript{\rm 2}CASIA 
    ~\textsuperscript{\rm 3}GigaAI
    ~\textsuperscript{\rm 4}Tsinghua University
    ~\textsuperscript{\rm 5}X-Humanoid
    ~\textsuperscript{\rm 6}FiveAges \\
    ~\texttt{yuan.xu@nlpr.ia.ac.cn}
    \\
    ~Project Page: \url{https://EgoDemoGen.github.io/}
}

\begin{document}
\maketitle

\begin{abstract}
Imitation learning based visuomotor policies have achieved strong performance in robotic manipulation, yet they often remain sensitive to egocentric viewpoint shifts. Unlike third-person viewpoint changes that only move the camera, egocentric shifts simultaneously alter both the camera pose and the robot action coordinate frame, making it necessary to jointly transfer action trajectories and synthesize corresponding observations under novel egocentric viewpoints. To address this challenge, we present EgoDemoGen, a framework that generates paired observation--action demonstrations under novel egocentric viewpoints through two key components: 1{)} EgoTrajTransfer, which transfers robot trajectories to the novel egocentric coordinate frame through motion-skill segmentation, geometry-aware transformation, and inverse kinematics filtering; and 2{)} EgoViewTransfer, a conditional video generation model that fuses a novel-viewpoint reprojected scene video and a robot motion video rendered from the transferred trajectory to synthesize photorealistic observations, trained with a self-supervised double reprojection strategy without requiring multi-viewpoint data. Experiments in simulation and real-world settings show that EgoDemoGen consistently improves policy success rates under both standard and novel egocentric viewpoints, with absolute gains of +24.6\% and +16.9\% in simulation and +16.0\% and +23.0\% on the real robot. Moreover, EgoViewTransfer achieves superior video generation quality for novel egocentric observations.
\end{abstract}

\section{Introduction}
\label{sec:introduction}

\begin{figure}[t]
    \centering
    \includegraphics[width=\columnwidth]{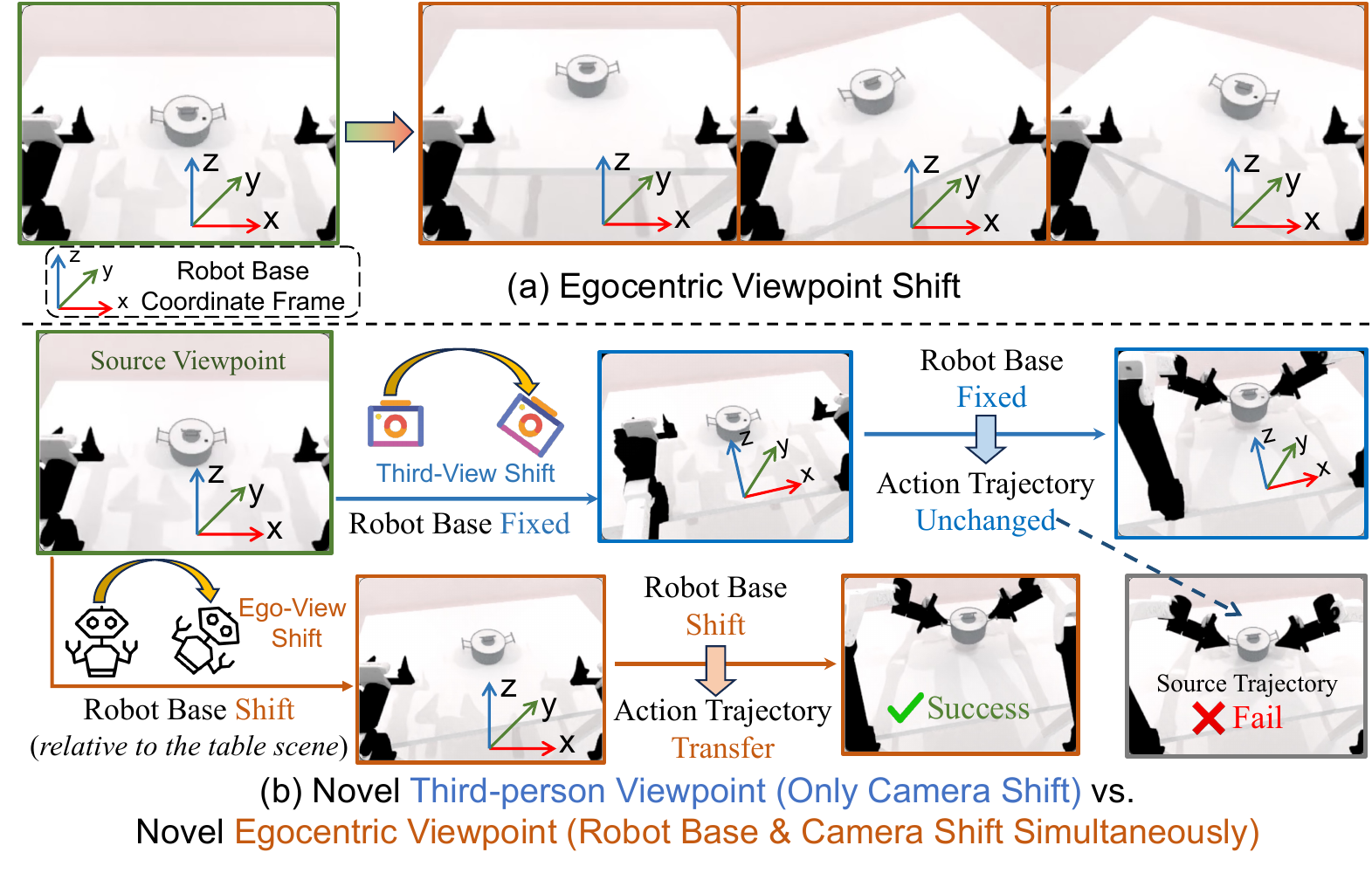}
    \caption{
        (a) Egocentric viewpoint shifts alter the robot base and the observed scene.
        (b) Third-person viewpoint changes only shift the camera (robot base fixed, action unchanged), whereas egocentric changes shift both the robot base and camera simultaneously; directly replaying the source trajectory under the new base frame fails, necessitating action trajectory transfer and observation synthesis.
    }
    \label{fig:intro}
\end{figure}

Imitation learning has become a dominant paradigm for robotic manipulation, with recent Vision-Language-Action models~\cite{chi2023diffusion, zhao2023learning, black2024pi0visionlanguageactionflowmodel, ghosh2024octo, liu2024rdt} achieving strong performance when trained on numerous demonstrations~\cite{wu2024robomind, khazatsky2024droid, walke2023bridgedata, o2024open}.
However, these policies remain sensitive to viewpoint shifts: policies trained from a single egocentric viewpoint often fail when the viewpoint changes during deployment~\cite{tian2025view, xing2025shortcut}, as illustrated in~\Cref{fig:intro}(a).
This is a practical concern because egocentric viewpoint shifts arise naturally from imprecise robot base positioning, platform reconfiguration, or environment layout changes.
Collecting demonstrations under every possible viewpoint is prohibitively expensive, motivating the generation of demonstrations under diverse egocentric viewpoints.

However, generating demonstrations under novel egocentric viewpoints is fundamentally different from synthesizing third-person novel views.
As shown in~\Cref{fig:intro}(b), a third-person viewpoint change only moves the camera, whereas an egocentric change shifts both the robot base and camera simultaneously, so the source action trajectory becomes invalid under the new base coordinate frame and must be transferred accordingly.
A valid novel egocentric demonstration therefore requires both a kinematically feasible action trajectory adapted to the new base frame and a visually realistic observation from the new viewpoint, and these two must be aligned.
Existing novel-view synthesis methods~\cite{sargent2024zeronvs, xue2025demogen, yang2025novel, tian2025view} generate observations without transferring actions, leading to visual--action misalignment under egocentric shifts.
Action-conditioned video generation and world models~\cite{wang2025embodiedreamer, rigter2024avid, bruce2024genie, luo2024grounding, hafner2025mastering} target prediction or planning rather than paired demonstration generation, and do not model the egocentric viewpoint changes.

This reveals a fundamental gap: no existing approach generates \emph{paired} observation--action demonstrations that remain consistent under egocentric viewpoint changes.
Closing this gap requires addressing two core challenges: (1) producing kinematically feasible robot actions that complete the task under the novel egocentric viewpoint, and (2) synthesizing realistic, temporally consistent observation videos that depict the transferred robot trajectories from the novel viewpoint.

To tackle these challenges, we propose \textbf{EgoDemoGen}, a framework for generating paired demonstrations from novel egocentric viewpoints.
On the action side, \textbf{EgoTrajTransfer} segments the source trajectory into motion and skill phases based on gripper states, applies geometry-aware transformations tailored to each phase, and reconstructs joint actions via inverse kinematics with feasibility filtering.
On the visual side, \textbf{EgoViewTransfer} synthesizes photorealistic observation videos by conditioning a video diffusion model on two decoupled inputs: a reprojected scene video encoding the novel viewpoint geometry, and a rendered robot motion video encoding the transferred trajectory, trained with a self-supervised double reprojection strategy without requiring multi-viewpoint data.
We evaluate EgoDemoGen in both simulation and real-world settings against five baselines spanning geometry-based and video-generation-based methods, and conduct analyses on ablation, data scaling, viewpoint generalization range, and base-camera decoupled generalization.

Our main contributions are as follows:
\begin{itemize}
    \item We present \textbf{EgoDemoGen}, a framework that generates paired observation--action demonstrations under novel egocentric viewpoints, improving policy generalization to egocentric viewpoint shifts.
    \item We propose \textbf{EgoTrajTransfer}, which transfers robot trajectories to the novel egocentric coordinate frame through motion-skill segmentation and geometry-aware transformation with inverse kinematics feasibility filtering.
    \item We propose \textbf{EgoViewTransfer}, a conditional video generation model that fuses reprojected scene video and rendered robot motion video to synthesize photorealistic observations from novel egocentric viewpoints, trained with a self-supervised double reprojection strategy.
    \item Experiments in simulation and on a real robot demonstrate that EgoDemoGen consistently improves policy success rates under both standard and novel egocentric viewpoints, while achieving higher video generation quality for novel egocentric observations.
\end{itemize}

% \begin{figure*}[t]
%     \centering
%     \includegraphics[width=\linewidth]{newfigures/overview.pdf}
%     \caption{
%         \textbf{Overview of EgoDemoGen.}
%         Given source demonstrations from a standard egocentric viewpoint, we generate novel demonstrations through four stages:
%         \textbf{(1)} sampling novel egocentric viewpoints via robot base motion;
%         \textbf{(2) EgoTrajTransfer} transfers trajectories to novel viewpoints through motion-skill segmentation and geometry-aware transformation, filtering out infeasible viewpoints via inverse kinematics;
%         \textbf{(3) EgoViewTransfer} synthesizes photorealistic observations by reprojecting the scene, rendering robot motion, and fusing both via conditional video generation;
%         \textbf{(4)} the generated demonstrations are combined with original data to train policies that generalize across egocentric viewpoints.
%     }
%     \label{fig:overview}
% \end{figure*}

\section{Related Work}
\label{sec:related_work}

\subsection{Data Generation for Policy Learning}
Existing demonstration generation methods fall into three streams.
(1)~\textbf{Geometry-based methods} leverage depth-based reprojection or 3D reconstruction to render novel-view observations but preserve original actions, causing visual--action mismatch under egocentric viewpoint shifts~\cite{xue2025demogen, yang2025novel, zhou2023nerf}.
(2)~\textbf{Visual synthesis methods} apply image or video generation models to produce novel-viewpoint observations.
VISTA~\cite{tian2025view} uses zero-shot novel view synthesis for third-person view augmentation, while TrajectoryCrafter~\cite{yu2025trajectorycrafter} redirects camera trajectories via diffusion models and Phantom~\cite{lepert2025phantom} inpaints missing regions after view transformation.
However, these methods either do not generate corresponding actions or do not account for action-frame changes under egocentric viewpoint shifts~\cite{sargent2024zeronvs, chen2023genaug, chen2025rovi}.
(3)~\textbf{Motion retargeting methods} adapt action trajectories to new object poses or embodiments but operate from a third-person viewpoint without addressing egocentric viewpoint changes~\cite{mandlekar2023mimicgen, ameperosa2025rocoda, lin2025constraint}.
EgoDemoGen differs from all three streams by jointly transferring actions to the novel egocentric frame and synthesizing corresponding observation videos, producing \emph{paired} demonstrations with visual--action alignment.

% \begin{figure*}[t]
%     \centering
%     \includegraphics[width=\linewidth]{figures/overview2.pdf}
%     \caption{
%         \textbf{Overview of EgoDemoGen.}
%         \textbf{(1) Egocentric View Transform:} a \emph{novel egocentric view} is specified by robot base motion $(\Delta x,\ \Delta y,\ \Delta \theta)$.
%         \textbf{(2) Action Retargeting:} the original joint actions $Q$ are \emph{retargeted} into the novel robot base frame to yield kinematically feasible joint actions $\tilde{Q}$.
%         \textbf{(3) Novel Egocentric Observations:} starting from the original observation video $V$, we mask the robot, reproject the scene to the novel viewpoint, perform hole filling, and apply \textbf{EgoViewTransfer} to synthesize coherent observations $\tilde{V}$.
%         \textbf{(4) Novel Demonstrations \& Policy Training:} we obtain aligned pairs $(\tilde{V},\ \tilde{Q})$ for training egocentric viewpoint-robust policies. 
%     }
%     \label{fig:overview}
% \end{figure*}

\begin{figure*}[t]
  \centering
  \includegraphics[width=\linewidth]{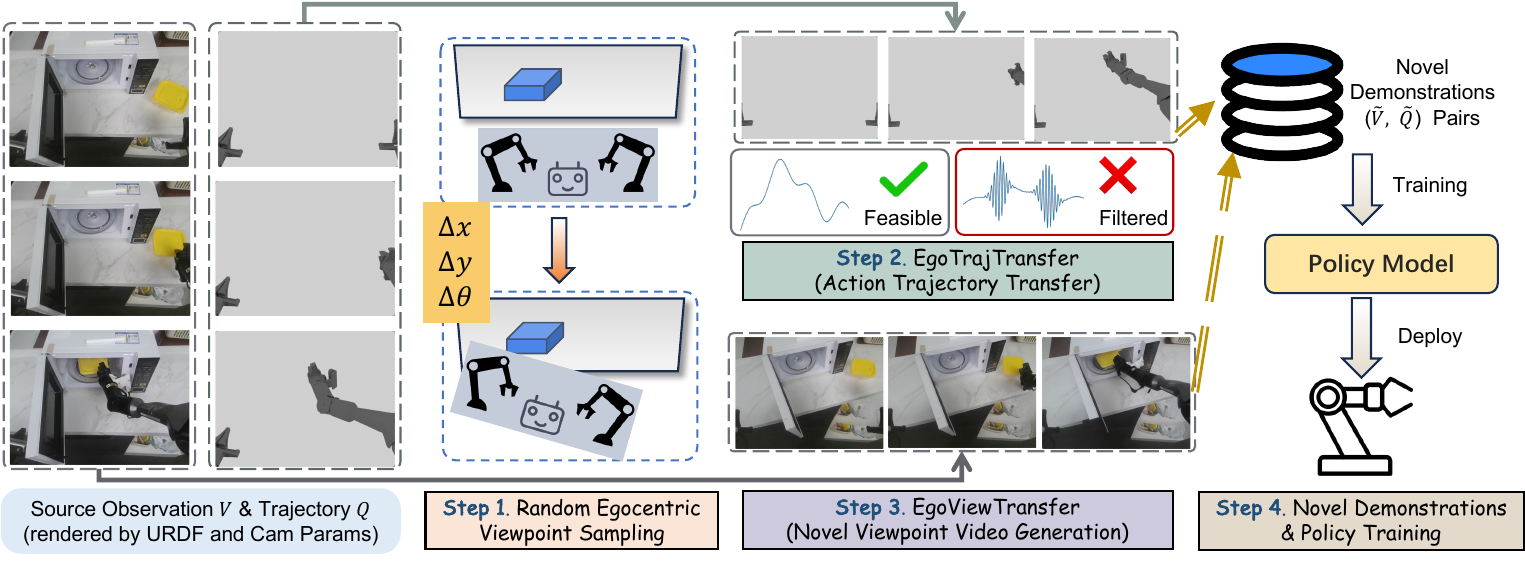}
  \caption{
      \textbf{Overview of EgoDemoGen.}
      Given source demonstrations from a standard egocentric viewpoint, we generate novel demonstrations through four steps:
      \textbf{(1)} sampling novel egocentric viewpoints via robot base motion;
      \textbf{(2) EgoTrajTransfer} produces kinematically feasible action trajectories adapted to the novel egocentric coordinate frame, filtering out infeasible viewpoints via inverse kinematics;
      \textbf{(3) EgoViewTransfer} synthesizes photorealistic observation videos from the novel egocentric viewpoint, depicting the transferred robot motion;
      \textbf{(4)} the generated demonstrations are combined with original data to train policies that generalize across egocentric viewpoints.
  }
  \label{fig:overview}
\end{figure*}

\subsection{Video Generation Models in Robotics}
Recent video generation models have been applied to robotics along three directions.
(1)~\textbf{Video-as-policy} approaches train or adapt a video generation model and decode executable actions from its rollouts, leveraging the visual prior learned from large-scale video data to improve policy generalization across tasks and environments~\cite{cheang2024gr, liang2025video, liang2025dreamitate}.
(2)~\textbf{Predict-then-act} methods first synthesize future observations, often conditioned on language instructions or goal images, and then infer actions or planning supervision from the generated sequences, decoupling visual prediction from motor control~\cite{du2023learning, luo2024grounding, patel2025robotic}.
(3)~\textbf{Action-conditioned world models} generate future video conditioned on candidate actions, serving as learned simulators for model-based planning or as data engines that produce additional training trajectories~\cite{zhu2024irasim, zhou2024robodreamer, liu2025robotransfer, jang2025dreamgen}.
While all three directions leverage video generation for robot learning, none of them explicitly model egocentric viewpoint shifts or generate paired observation--action demonstrations under novel viewpoints.
EgoDemoGen instead finetunes a video generation model to synthesize observation videos aligned with transferred actions in the novel egocentric frame, targeting paired demonstration generation rather than planning or simulation.

\section{Method}
\label{sec:method}

% \begin{figure*}[t]
%   \centering
%   \includegraphics[width=\linewidth]{newfigures/overview-3.pdf}
%   \caption{
%       \textbf{Overview of EgoDemoGen.}
%       Given source demonstrations from a standard egocentric viewpoint, we generate novel demonstrations through four steps:
%       \textbf{(1)} sampling novel egocentric viewpoints via robot base motion;
%       \textbf{(2) EgoTrajTransfer} produces kinematically feasible action trajectories adapted to the novel egocentric coordinate frame, filtering out infeasible viewpoints via inverse kinematics;
%       \textbf{(3) EgoViewTransfer} synthesizes photorealistic observation videos from the novel egocentric viewpoint, depicting the transferred robot motion;
%       \textbf{(4)} the generated demonstrations are combined with original data to train policies that generalize across egocentric viewpoints.
%   }
%   \label{fig:overview}
% \end{figure*}

Our method, \textbf{EgoDemoGen}, generates novel egocentric demonstrations from a single source demonstration by transferring both robot trajectories and visual observations to new viewpoints.
As illustrated in \Cref{fig:overview}, we sample novel egocentric viewpoints and introduce two main components: \textbf{EgoTrajTransfer} (\Cref{sec:trajectory_transfer}) transfers trajectories with viewpoint feasibility filtering, and \textbf{EgoViewTransfer} (\Cref{sec:view_synthesis}) synthesizes observations via conditional video generation.
We train EgoViewTransfer using a self-supervised double reprojection strategy (\Cref{sec:training}).

\subsection{Problem Formulation}
\label{sec:problem_formulation}

Let $\mathcal{D} = \{(V^{(i)}, Q^{(i)})\}_{i=1}^N$ be demonstrations from egocentric viewpoint $v_{\text{src}}$, where $V^{(i)} = \{I_t\}_{t=1}^T$ denotes RGB video and $Q^{(i)} = \{q_t\}_{t=1}^T$ the joint trajectory with $q_t \in \mathbb{R}^{d}$ (dimension $d$ depends on robot configuration).

\textbf{Egocentric viewpoint changes.}
Unlike third-person views, egocentric viewpoint shifts correspond to robot base transformations in world coordinates.
Since the camera is head-mounted, base motion directly induces viewpoint changes.
For practical mobile manipulation scenarios, we focus on planar motion parameterized by $v = (\Delta x, \Delta y, \Delta \theta)$, inducing coordinate transformation $\Delta T \in \text{SE}(3)$.

\textbf{Objective.}
Given a source demonstration $(V, Q, v_{\text{src}})$ and a target viewpoint $v_{\text{nov}}$, our goal is to generate a novel demonstration $(\tilde{V}, \tilde{Q}, v_{\text{nov}})$ that executes the same task from the new viewpoint.
This requires two mappings:
\begin{equation}
    \tilde{Q} = \mathcal{T}(Q; \Delta T), \quad
    \tilde{V} = \mathcal{G}(V, \tilde{Q}; \Delta T),
    \label{eq:objective}
\end{equation}
where $\mathcal{T}$ is a trajectory transfer operator that produces kinematically feasible joint trajectories adapted to the novel viewpoint, and $\mathcal{G}$ is a conditional video generator that synthesizes photorealistic observations aligned with the transferred trajectory.
The key challenge is to ensure visual-action alignment: the robot motion depicted in $\tilde{V}$ must correspond precisely to the actions in $\tilde{Q}$, enabling effective policy learning from generated demonstrations.

\subsection{EgoTrajTransfer}
\label{sec:trajectory_transfer}

% \begin{figure}[t]
%     \centering
%     % \fbox{\rule{0pt}{2in}\rule{\columnwidth}{0pt}}
%     \includegraphics[width=\linewidth]{newfigures/EgoTrajTransfer-3.pdf}
%     \caption{
%         \textbf{EgoTrajTransfer pipeline.}
%         Top: source trajectory segmented into motion segments (free-space movement) and skill segments (contact-rich manipulation) by gripper states.
%         Bottom: transferred trajectory using position scaling and orientation interpolation for motion segments, rigid transformation for skill segments.
%     }
%     \label{fig:trajectory_transfer}
% \end{figure}

% \begin{figure*}[t]
%     \centering
%     % \fbox{\rule{0pt}{2in}\rule{\linewidth}{0pt}}
%     \includegraphics[width=\linewidth]{newfigures/EgoViewTransfer.pdf}
%     \caption{
%         \textbf{EgoViewTransfer pipeline.}
%         We synthesize novel-viewpoint observations through three stages.
%         First, scene video preparation: reproject the original video to the novel viewpoint, mask the robot region, and inpaint to obtain clean background.
%         Second, robot motion rendering: render robot motion from the transferred trajectory using URDF and camera parameters.
%         Third, conditional video generation: fuse both videos via DiT-based diffusion model with dual-video conditioning.
%     }
%     \label{fig:view_synthesis}
% \end{figure*}

\begin{figure}[t]
    \centering
    % \fbox{\rule{0pt}{2in}\rule{\columnwidth}{0pt}}
    \includegraphics[width=\linewidth]{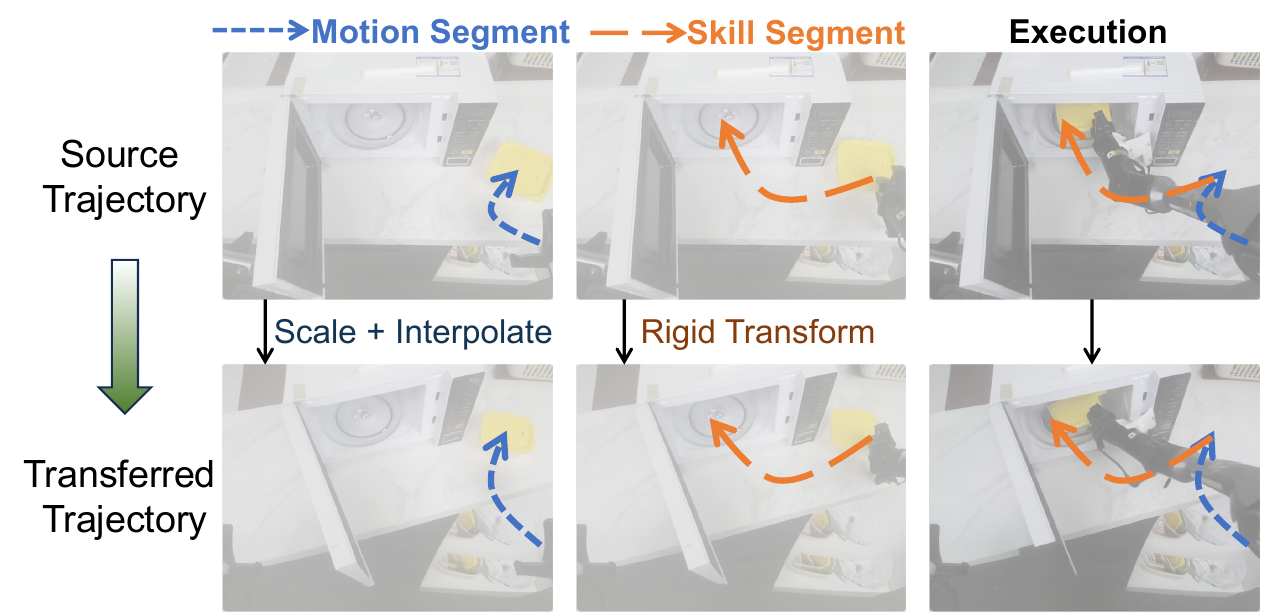}
    \caption{
        \textbf{EgoTrajTransfer pipeline.}
        Top: source trajectory segmented into motion segments (free-space movement) and skill segments (contact-rich manipulation) by gripper states.
        Bottom: transferred trajectory using position scaling and orientation interpolation for motion segments, rigid transformation for skill segments.
    }
    \label{fig:trajectory_transfer}
\end{figure}

We propose \textbf{EgoTrajTransfer}, which adapts robot trajectories to novel viewpoints while preserving task semantics.
% The key insight is that free-space motion segments and contact-rich skill segments require different transformations: motion segments have start and end poses that shift under the new viewpoint, so their path must be rescaled and reoriented, whereas skill segments maintain fixed end-effector poses relative to the scene and can be directly rigid-transformed.
As shown in \Cref{fig:trajectory_transfer}, we first segment trajectories based on gripper states into motion and skill segments, then apply tailored geometry-aware transformations to each segment type, and finally reconstruct joint angles via inverse kinematics.
The complete algorithm is presented in \Cref{alg:trajectory_transfer}.

\textbf{Trajectory segmentation.}
We segment each arm independently based on gripper state $g_t^a$ for arm $a \in \{L, R\}$: closed gripper ($g_t^a < \tau_{\text{close}}$) indicates contact-rich \emph{skill} segments, while open gripper indicates free-space \emph{motion} segments.
To handle sensor noise, we synchronize segment boundaries between arms within temporal window $w_{\text{sync}}$.

\textbf{Motion segment transfer.}
For motion segments, we preserve end-effector path structure while adapting to new endpoints.
We compute the new end pose via $T_{\text{end}}^{\text{new}} = \Delta T \cdot T_{\text{end}}^{\text{old}}$, then interpolate positions and orientations:
\begin{equation}
    p^{\text{new}}(t) = R_{\text{align}} (p^{\text{old}}(t) - p_{\text{start}}^{\text{old}}) s + p_{\text{start}}^{\text{new}},
    \label{eq:motion_position}
\end{equation}
where $s = \|p_{\text{end}}^{\text{new}} - p_{\text{start}}^{\text{new}}\| / \|p_{\text{end}}^{\text{old}} - p_{\text{start}}^{\text{old}}\|$ and $R_{\text{align}}$ aligns path directions.

\textbf{Skill segment transfer.}
For skill segments, we apply rigid body transformation to preserve object-relative motion.
The coordinate transformation $\Delta T$ induced by the base motion is directly applied to all end-effector poses:
\begin{equation}
    T_e^{\text{new}}(t) = \Delta T \cdot T_e^{\text{old}}(t).
    \label{eq:skill_transform}
\end{equation}

\textbf{Inverse kinematics and filtering.}
We solve batch IK using CuRobo~\cite{sundaralingam2023curobo} for each arm independently, using original joint angles as seeds to preserve arm configuration.
Failed frames are interpolated from nearest successful frames.
We filter infeasible viewpoints based on IK success rate exceeding $\tau_{\text{IK}}$ and maximum joint angle change below $\tau_{\text{jump}}$.

\begin{algorithm}[t]
\caption{EgoTrajTransfer}
\label{alg:trajectory_transfer}
\begin{algorithmic}[1]
\Require Source trajectory $Q = \{q_t\}_{t=1}^T$, transformation $\Delta T$
\Ensure Transferred trajectory $\tilde{Q} = \{\tilde{q}_t\}_{t=1}^T$

\State \textbf{// Step 1: Segment dual-arm trajectory}
\For{arm $a \in \{L, R\}$}
    \State Segment into motion/skill based on gripper state $g_t^a$
\EndFor
\State Synchronize segment boundaries between arms (window $w_{\text{sync}}$)

\State \textbf{// Step 2: Transfer segments}
\For{each segment $s$ of each arm}
    \State $T_e^a(t) = \text{FK}_a(q_t^a)$ \Comment{Forward kinematics}
    \If{$s$ is motion segment}
        \State $T_{\text{end}}^{\text{new}} = \Delta T \cdot T_{\text{end}}^{\text{old}}$; compute scale $s$ and $R_{\text{align}}$
        \State $p^{\text{new}}(t) \!=\! R_{\text{align}} (p^{\text{old}}(t) - p_{\text{start}}^{\text{old}}) s + p_{\text{start}}^{\text{new}}$ \Comment{Scale \& Align}
        \State $R^{\text{new}}(t) = \text{Slerp}(R_{\text{start}}^{\text{new}}, R_{\text{end}}^{\text{new}}; \alpha(t))$ \Comment{Interpolate}
    \ElsIf{$s$ is skill segment}
        \State $T_e^{\text{new}}(t) = \Delta T \cdot T_e^{\text{old}}(t)$ \Comment{Rigid Transform}
    \EndIf
\EndFor

\State \textbf{// Step 3: Solve IK and filter}
\For{arm $a \in \{L, R\}$}
    \State $\tilde{q}_t^a = \text{IK}_a(T_e^{a,\text{new}}(t))$ with seed $q_t^a$; interpolate failures
\EndFor
\If{IK success rate $< \tau_{\text{IK}}$ \textbf{or} max joint change $> \tau_{\text{jump}}$}
    \State \Return \texttt{INFEASIBLE}
\EndIf
\State \Return $\tilde{Q} = \{[\tilde{q}_t^L; \tilde{q}_t^R]\}_{t=1}^T$
\end{algorithmic}
\end{algorithm}

% \begin{figure*}[t]
%   \centering
%   % \fbox{\rule{0pt}{2in}\rule{\linewidth}{0pt}}
%   \includegraphics[width=\linewidth]{newfigures/EgoViewTransfer.pdf}
%   \caption{
%       \textbf{EgoViewTransfer pipeline.}
%       We synthesize novel-viewpoint observations through three stages.
%       First, scene video preparation: reproject the original video to the novel viewpoint, mask the robot region, and inpaint to obtain clean background.
%       Second, robot motion rendering: render robot motion from the transferred trajectory using URDF and camera parameters.
%       Third, conditional video generation: fuse both videos via DiT-based diffusion model with dual-video conditioning, where DiT is trained and VAE remains frozen.
%   }
%   \label{fig:view_synthesis}
% \end{figure*}

\subsection{EgoViewTransfer}
\label{sec:view_synthesis}

\begin{figure*}[t]
    \centering
    % \fbox{\rule{0pt}{2in}\rule{\linewidth}{0pt}}
    \includegraphics[width=\linewidth]{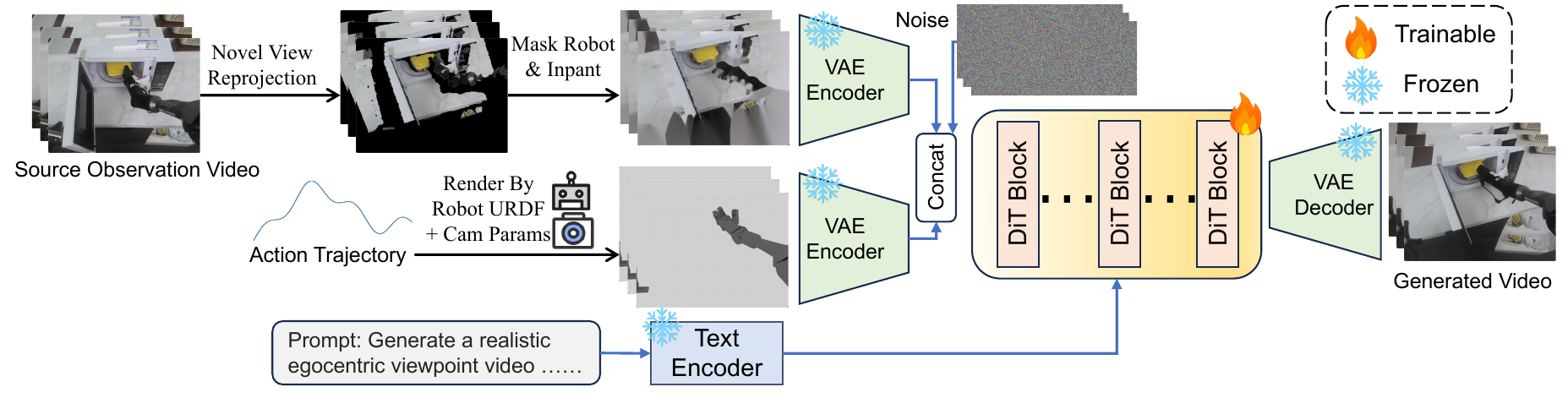}
    \caption{
        \textbf{EgoViewTransfer pipeline.}
        We synthesize novel-viewpoint observations through three stages.
        First, scene video preparation: reproject the original video to the novel viewpoint, mask the robot region, and inpaint to obtain clean background.
        Second, robot motion rendering: render robot motion from the transferred trajectory using URDF and camera parameters.
        Third, conditional video generation: fuse both videos via DiT-based diffusion model with dual-video conditioning.
    }
    \label{fig:view_synthesis}
\end{figure*}

We propose \textbf{EgoViewTransfer}, a video diffusion model that synthesizes photorealistic RGB observations $\tilde{V}$ from the novel viewpoint, conditioned on the transferred trajectory $\tilde{Q}$.
As shown in \Cref{fig:view_synthesis}, the pipeline consists of three stages: scene video preparation, robot motion rendering, and conditional video generation with a unified DiT-based diffusion model.

\textbf{Scene video preparation.}
To obtain a clean scene background from the novel viewpoint, we first reproject each frame of the original video $V$ to the novel viewpoint, producing $\hat{I}_t^{\text{nov}}$ with holes and artifacts.
We then remove the robot region by rendering robot mask $M_t$ using the original trajectory $Q$, URDF file, and novel viewpoint parameters, and applying the mask: $I_t^{\text{masked}} = (1 - M_t) \odot \hat{I}_t^{\text{nov}}$.
Finally, we inpaint the masked regions using the classic method INPAINT\_TELEA~\cite{telea2004image} to obtain scene video $V_S^{\text{nov}}$.

\textbf{Robot motion rendering.}
We render a robot-only video $V_R^{\text{nov}}$ that depicts the robot motion in the novel viewpoint using the transferred trajectory $\tilde{Q}$, URDF file, and novel viewpoint camera parameters (intrinsics and extrinsics).
This provides explicit robot motion conditioning for the video generator.

\textbf{Conditional video generation.}
We generate the final video using EgoViewTransfer, a video diffusion model that takes both scene and robot videos as conditions:
\begin{equation}
    \tilde{V} = \text{EgoViewTransfer}(V_S^{\text{nov}}, V_R^{\text{nov}}; \theta).
    \label{eq:video_generation}
\end{equation}
The scene video and robot video serve as decoupled conditions: the scene video encodes the novel viewpoint transformation, while the robot video encodes the transferred trajectory.
This decoupling enables the model to generalize to arbitrary combinations of viewpoints and trajectories at inference time.

\subsection{Self-Supervised Training}
\label{sec:training}

\begin{figure}[t]
    \centering
    % \fbox{\rule{0pt}{1.5in}\rule{\columnwidth}{0pt}}
    \includegraphics[width=\linewidth]{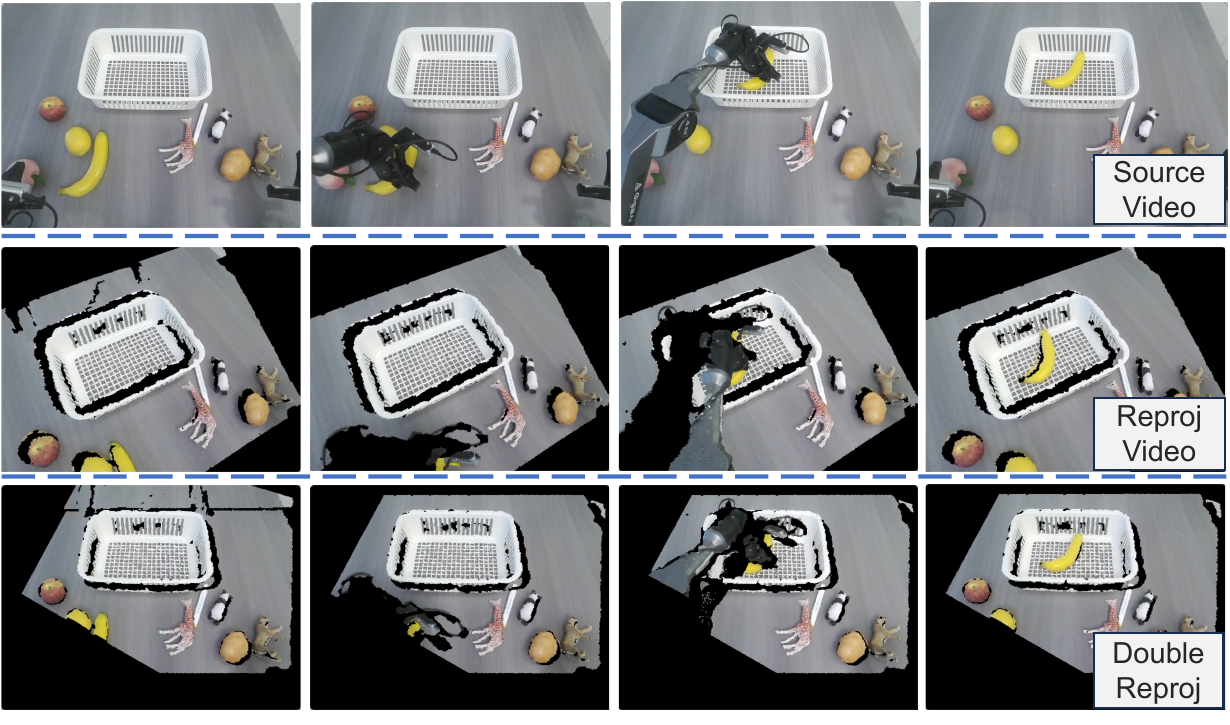}
    \caption{
        \textbf{Double reprojection training strategy.}
        Top: source video serves as ground truth.
        Bottom: double reprojection video serves as training input, constructed by reprojecting the source video to a random novel viewpoint and back to the source viewpoint, which introduces reprojection artifacts and missing regions that the model learns to repair.
        This self-supervised approach enables training without multi-viewpoint demonstrations.
    }
    \label{fig:double_reprojection}
\end{figure}

Training EgoViewTransfer requires paired data $(V_S^{\text{nov}}, V_R^{\text{nov}}, V^{\text{target}})$, but we only have demonstrations from a single viewpoint.
Thus we enable self-supervised training through a double reprojection strategy, as illustrated in \Cref{fig:double_reprojection}, which creates training pairs from the source viewpoint data without requiring multi-viewpoint demonstrations.

\textbf{Double reprojection.}
For each source demonstration $(V, Q, v_{\text{src}})$, we sample a random training viewpoint $v_{\text{train}}$ and construct a training pair through double reprojection:
(1) \textbf{Scene video}: reproject $V$ from $v_{\text{src}}$ to $v_{\text{train}}$, then reproject back to $v_{\text{src}}$, mask robot, and inpaint to obtain $V_S^{\text{train}}$. This double reprojection introduces artifacts and missing regions that simulate viewpoint transformation challenges;
(2) \textbf{Robot video}: render robot motion using the \emph{original} trajectory $Q$ at $v_{\text{src}}$ to obtain $V_R^{\text{train}}$;
(3) \textbf{Target}: use the source video $V$ as ground truth.
The model learns to repair reprojection artifacts and composite robot motion, which generalizes to novel viewpoints with transferred trajectories at inference time.

\textbf{Training objective.}
We train EgoViewTransfer using the standard diffusion denoising objective:
\begin{equation}
    \mathcal{L} = \mathbb{E}_{t, \epsilon} \left[ \| \epsilon - \epsilon_\theta(z_t, V_S^{\text{train}}, V_R^{\text{train}}, t) \|^2 \right],
    \label{eq:training_loss}
\end{equation}
where $z_t$ is the noisy latent at diffusion timestep $t$, $\epsilon \sim \mathcal{N}(0, I)$ is the noise, and $\epsilon_\theta$ is the denoising network.

% \textbf{Decoupled control.}
% As described in \Cref{sec:view_synthesis}, the decoupled conditioning enables generalization to any $(\tilde{Q}, v_{\text{nov}})$ combination at inference.

\section{Experiments}
\label{sec:exp}

We conduct experiments to answer the following questions:
\begin{itemize}
    \item \textbf{Q1:} Does EgoDemoGen improve policy robustness under novel egocentric viewpoints in both simulation and real-world settings? (Sec.~\ref{subsec:main_results})
    \item \noindent \textbf{Q2:} Are all components—double reprojection, mask \& inpaint, action conditioning, and action trajectory transfer—necessary for the performance gains? (Sec.~\ref{subsec:ablation})
    \item \textbf{Q3:} What is the optimal ratio between real and synthetic demonstrations, and how does increasing synthetic data affect performance? (Sec.~\ref{subsec:data_scaling})
    \item \textbf{Q4:} How does the egocentric viewpoint range of generated demonstrations affect generalization to unseen viewpoints? (Sec.~\ref{subsec:viewpoint_generalization})
    \item \textbf{Q5:} Can EgoDemoGen handle decoupled robot base and egocentric camera motion? (Sec.~\ref{subsec:decoupled_motion})
\end{itemize}

% \begin{figure}[t]
%     \centering
%     \includegraphics[width=\columnwidth]{newfigures/task-3.pdf}
%     \caption{
%         \textbf{Real-world tasks and viewpoint configurations.}
%         Top row: the five real-world tasks on the Mobile ALOHA platform.
%         Bottom row: egocentric viewpoint configurations used for evaluation—one standard viewpoint (leftmost) and four novel viewpoints.
%     }
%     \label{fig:tasks}
% \end{figure}

\subsection{Experimental Setup}
\label{subsec:setup}

\begin{figure}[t]
    \centering
    \includegraphics[width=\columnwidth]{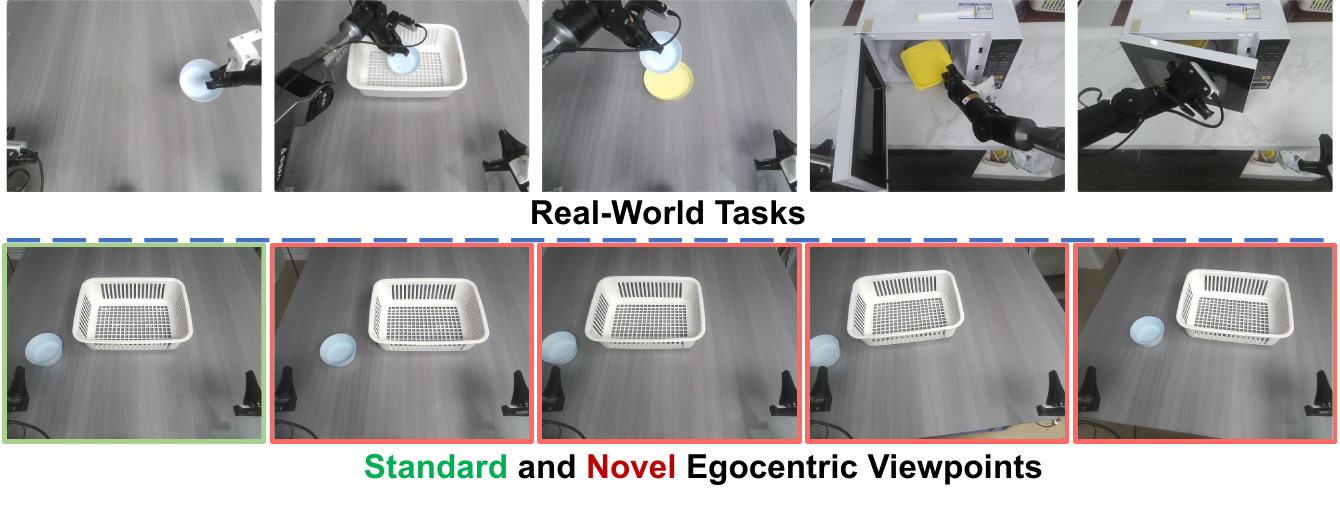}
    \caption{
        \textbf{Real-world tasks and viewpoint configurations.}
        Top row: the five real-world tasks on the Mobile ALOHA platform.
        Bottom row: egocentric viewpoint configurations used for evaluation—one standard viewpoint (leftmost) and four novel viewpoints.
    }
    \label{fig:tasks}
\end{figure}

\begin{table*}[t]
    \centering
    \caption{\textbf{Simulation results: task success rates (\%) on seven RoboTwin 2.0 tasks.}
    Std.: standard egocentric viewpoint; Nov.: novel egocentric viewpoint (100 trials each).
    Best results per column in \textbf{bold}.}
    \label{tab:sim_main}
    \resizebox{0.95\textwidth}{!}{%
    \begin{tabular}{l cc cc cc cc cc cc cc cc}
        \toprule
        & \multicolumn{2}{c}{\textit{Adjust}} 
        & \multicolumn{2}{c}{\textit{Beat}} 
        & \multicolumn{2}{c}{\textit{Lift}} 
        & \multicolumn{2}{c}{\textit{Handover}} 
        & \multicolumn{2}{c}{\textit{Open}} 
        & \multicolumn{2}{c}{\textit{Place}} 
        & \multicolumn{2}{c}{\textit{Put}} 
        & \multicolumn{2}{c}{\textbf{Average}} \\
        \cmidrule(lr){2-3}
        \cmidrule(lr){4-5}
        \cmidrule(lr){6-7}
        \cmidrule(lr){8-9}
        \cmidrule(lr){10-11}
        \cmidrule(lr){12-13}
        \cmidrule(lr){14-15}
        \cmidrule(lr){16-17}
        \textbf{Method} 
        & Std. & Nov. 
        & Std. & Nov. 
        & Std. & Nov. 
        & Std. & Nov. 
        & Std. & Nov. 
        & Std. & Nov. 
        & Std. & Nov. 
        & Std. & Nov. \\
        \midrule
        Standard Viewpoint 
        & 92 & 38 & 13 & 3 & 33 & 8 & 1 & 1 & 38 & 25 & 26 & 2 & 0 & 0 & 29.0 & 11.0 \\
        Direct Reprojection 
        & 95 & 50 & 40 & 10 & 31 & 14 & 12 & 5 & 69 & 50 & 51 & 12 & 15 & 2 & 44.7 & 20.4 \\
        TrajectoryCrafter~\cite{yu2025trajectorycrafter} 
        & 96 & 46 & 13 & 7 & 37 & 14 & 44 & 19 & 63 & 39 & 51 & 7 & 11 & 2 & 45.0 & 19.1 \\
        Phantom~\cite{lepert2025phantom} 
        & 96 & 52 & 17 & 11 & 37 & \textbf{18} & 5 & 3 & 69 & 48 & 50 & 8 & 2 & 2 & 39.4 & 20.3 \\
        VISTA~\cite{tian2025view} 
        & \textbf{100} & 51 & \textbf{45} & 15 & \textbf{49} & 11 & 34 & 10 & \textbf{73} & \textbf{54} & 54 & 5 & \textbf{20} & 4 & \textbf{53.6} & 21.4 \\
        \rowcolor{gray!20}
        \textbf{EgoDemoGen} 
        & 99 & \textbf{61} & 41 & \textbf{21} & 45 & 15 & \textbf{56} & \textbf{20} & 66 & 46 & \textbf{60} & \textbf{25} & 8 & \textbf{7} & \textbf{53.6} & \textbf{27.9} \\
        \bottomrule
    \end{tabular}%
    }
  \end{table*}
  
  \begin{table}[t]
    \centering
    \caption{\textbf{Real-world results: task success rates (\%) on five Mobile ALOHA tasks.}
    S: standard egocentric viewpoint; N: novel egocentric viewpoints averaged over four fixed viewpoints (20 trials each).
    Best per column in \textbf{bold}.}
    \label{tab:real_main}
    \small
    \setlength{\tabcolsep}{2.5pt}
    \resizebox{\columnwidth}{!}{%
    \begin{tabular}{l cc cc cc cc cc cc}
        \toprule
        & \multicolumn{2}{c}{\makecell{\textit{Pick Up}\\\textit{Bowl}}} 
        & \multicolumn{2}{c}{\makecell{\textit{Pl. Bowl}\\\textit{Basket}}} 
        & \multicolumn{2}{c}{\makecell{\textit{Pl. Bowl}\\\textit{Plate}}} 
        & \multicolumn{2}{c}{\makecell{\textit{Pl. Box}\\\textit{Micro.}}} 
        & \multicolumn{2}{c}{\makecell{\textit{Close}\\\textit{Micro.}}} 
        & \multicolumn{2}{c}{\textbf{Avg.}} \\
        \cmidrule(lr){2-3}
        \cmidrule(lr){4-5}
        \cmidrule(lr){6-7}
        \cmidrule(lr){8-9}
        \cmidrule(lr){10-11}
        \cmidrule(lr){12-13}
        & S & N & S & N & S & N & S & N & S & N & S & N \\
        \midrule
        Std.\ View 
        & 70 & 45 & 65 & 40 & 45 & 25 & 50 & 20 & 70 & 55 & 60.0 & 37.0 \\
        Direct Reproj. 
        & 75 & 30 & 60 & 70 & 55 & 30 & 55 & 40 & 75 & 65 & 64.0 & 47.0 \\
        \rowcolor{gray!20}
        \textbf{EgoDemoGen} 
        & \textbf{90} & \textbf{70} & \textbf{80} & \textbf{75} & \textbf{65} & \textbf{40} & \textbf{65} & \textbf{45} & \textbf{80} & \textbf{70} & \textbf{76.0} & \textbf{60.0} \\
        \bottomrule
    \end{tabular}%
    }
  \end{table}
  
  \begin{table}[t]
    \centering
    \caption{\textbf{Video generation quality.}
    Simulation: compared against ground-truth novel-viewpoint videos.
    Real world: evaluated under the double-reprojection setting (standard viewpoint).
    Best results per section in \textbf{bold}.}
    \label{tab:video_quality}
    \small
    \setlength{\tabcolsep}{4pt}
    \resizebox{0.95\columnwidth}{!}{%
    \begin{tabular}{lcccc}
        \toprule
        \textbf{Method} & \textbf{PSNR}$\uparrow$ & \textbf{SSIM}$\uparrow$ & \textbf{LPIPS}$\downarrow$ & \textbf{FVD}$\downarrow$ \\
        \midrule
        \multicolumn{5}{l}{\textit{Simulation (novel egocentric viewpoint)}} \\
        \addlinespace[2pt]
        Direct Reprojection & 10.10 & 0.711 & 0.335 & 621.5 \\
        TrajectoryCrafter~\cite{yu2025trajectorycrafter} & 12.46 & 0.728 & 0.264 & 483.1 \\
        VISTA~\cite{tian2025view} & 12.67 & 0.687 & 0.271 & 1451.3 \\
        Phantom~\cite{lepert2025phantom} & 16.97 & 0.786 & 0.174 & 1131.0 \\
        \rowcolor{gray!20}
        \textbf{EgoViewTransfer} & \textbf{26.03} & \textbf{0.886} & \textbf{0.081} & \textbf{133.5} \\
        \midrule
        \multicolumn{5}{l}{\textit{Real world (double reprojection, standard viewpoint)}} \\
        \addlinespace[2pt]
        Direct Reprojection & 10.35 & 0.546 & 0.496 & 968.7 \\
        \rowcolor{gray!20}
        \textbf{EgoViewTransfer} & \textbf{26.93} & \textbf{0.890} & \textbf{0.087} & \textbf{148.6} \\
        \bottomrule
    \end{tabular}%
    }
  \end{table}

\begin{figure*}[t]
    \centering
    \includegraphics[width=\textwidth]{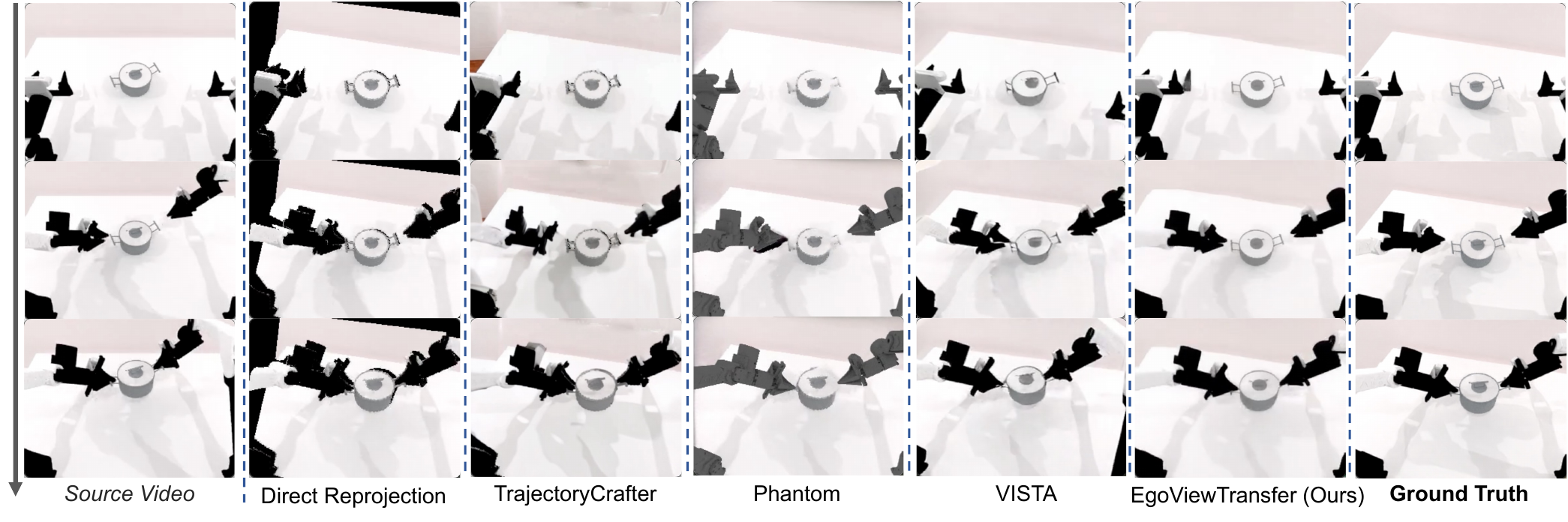}
    \caption{
        \textbf{Qualitative comparison of novel-viewpoint video generation in simulation.}
        From left to right: Source Video (standard egocentric viewpoint), Direct Reprojection, TrajectoryCrafter~\cite{yu2025trajectorycrafter}, Phantom~\cite{lepert2025phantom}, VISTA~\cite{tian2025view}, EgoViewTransfer (Ours), and Ground Truth.
        The ground truth is obtained by replaying the transferred trajectory in simulation and rendering from the novel egocentric viewpoint.
    }
    \label{fig:video_comparison}
\end{figure*}

We evaluate EgoDemoGen in both simulation and real-world settings.

\textbf{Simulation.}
We use the RoboTwin 2.0~\cite{chen2025robotwin} environment with dual ARX-X5 manipulators and a head-mounted egocentric camera.
We evaluate on seven tasks: \textit{Adjust Bottle}, \textit{Beat Block Hammer}, \textit{Lift Pot}, \textit{Handover Block}, \textit{Open Laptop}, \textit{Place Burger Fries}, and \textit{Put Object Cabinet}.
For each task, 25 demonstrations are collected from the standard egocentric viewpoint via scripted policies.
Each method is evaluated over 100 trials per task for both the standard egocentric viewpoint and novel egocentric viewpoints sampled within $\Delta x \in [-0.1, 0.1]$\,m, $\Delta y \in [-0.1, 0.1]$\,m, $\Delta \theta \in [-10, 10]$\,degrees.

\textbf{Real world.}
We use a Mobile ALOHA platform with dual arms and a head-mounted egocentric camera.
We evaluate on five tasks: \textit{Pick Up Bowl}, \textit{Place Bowl Basket}, \textit{Place Bowl Plate}, \textit{Place Box Microwave}, and \textit{Close Microwave}.
For each task, 50 demonstrations are collected from the standard egocentric viewpoint via teleoperation.
Each method is evaluated over 20 trials per task for the standard egocentric viewpoint and four fixed novel egocentric viewpoints, as shown in \Cref{fig:tasks}.

\textbf{Baselines.}
We compare EgoDemoGen with the following baselines: All of them that involve novel-viewpoint data use transferred action trajectory from EgoTrajTransfer (Sec.~\ref{sec:trajectory_transfer}), except
VISTA which keeps original actions as it targets third-person view
augmentation.
\begin{itemize}
    \item \textbf{Standard Viewpoint}: trained on standard egocentric viewpoint demonstrations only.
    \item \textbf{Direct Reprojection}: reprojecting original RGB-D to the novel viewpoint via known camera parameters.
    \item \textbf{TrajectoryCrafter}~\cite{yu2025trajectorycrafter}: diffusion-based camera trajectory redirection from original video and target extrinsics. (Simulation only.)
    \item \textbf{Phantom}~\cite{lepert2025phantom}: inpainting missing regions in the reprojected video and compositing the rendered robot video. (Simulation only.)
    \item \textbf{VISTA}~\cite{tian2025view}: third-person view augmentation via zero-shot novel view synthesis; actions unchanged. (Simulation only.)
\end{itemize}
Each baseline generates one novel-viewpoint demonstration per original, matching EgoDemoGen's 1:1 ratio.

\textbf{Implementation and Metrics.}
EgoViewTransfer is built on CogVideoX-5B-I2V~\cite{yang2024cogvideox}, with input channels expanded to 48 for dual-video conditioning (scene video and robot video).
The model is finetuned on standard egocentric-viewpoint demonstrations using the double reprojection strategy described in Sec.~\ref{sec:training}: 500 episodes across 50 tasks in simulation and 600 teleoperated episodes in the real world.
We use ACT~\cite{zhao2023learning} as the policy model in simulation and $\pi_0$~\cite{black2024pi0visionlanguageactionflowmodel} on the real robot.
Policy performance is measured by task success rate under the standard and novel egocentric viewpoints.
Video generation quality is assessed by PSNR, SSIM, LPIPS~\cite{zhang2018unreasonable}, and FVD~\cite{unterthiner2018towards}.
Training hyperparameters and additional details are provided in the supplementary material.

\subsection{Main Results (\textbf{Q1})}
\label{subsec:main_results}

\noindent\textbf{Simulation.}
\Cref{tab:sim_main} reports success rates across seven simulated tasks.
Training with only standard-viewpoint demonstrations suffers severe novel-view degradation, confirming that single-viewpoint data does not generalize to egocentric viewpoint shifts.
Among augmentation methods, VISTA~\cite{tian2025view} achieves the highest standard-view score yet limited novel-view gain because it synthesizes third-person views without egocentric action trajectory transfer, so novel-view policies still lack paired action supervision.
\textbf{EgoDemoGen} achieves the best novel-view performance on 6 out of 7 tasks while maintaining competitive standard-view accuracy, confirming that \emph{paired}, action-consistent egocentric demonstrations are more effective than view synthesis alone.

\noindent\textbf{Real world.}
\Cref{tab:real_main} confirms these findings on the real robot.
Direct Reprojection yields moderate gains over the standard-view baseline, but its performance is bounded by persistent reprojection artifacts that confuse the policy under large viewpoint shifts.
\textbf{EgoDemoGen} achieves substantially larger improvements (+16.0\% and +23.0\% on standard and novel views respectively), with consistent gains across all five tasks, demonstrating that high-quality generated demonstrations transfer effectively to real-world policy learning.

\noindent\textbf{Video generation quality.}
EgoViewTransfer outperforms all baselines across all four metrics in both simulation and real-world settings (\Cref{tab:video_quality}), indicating higher visual fidelity and temporal coherence.
As shown in \Cref{fig:video_comparison}, EgoViewTransfer effectively repairs the reprojection artifacts and boundary holes while producing robot motion consistent with the transferred trajectory.
In contrast, Direct Reprojection suffers from artifacts and black regions at view boundaries, TrajectoryCrafter and VISTA exhibit distorted robot arms and objects with temporal inconsistencies, and Phantom introduces noticeable appearance discrepancies in the composited robot region.
These results verify that the decoupled scene-robot conditioning design enables EgoViewTransfer to faithfully synthesize novel egocentric observations.

\subsection{Ablation Study (\textbf{Q2})}
\label{subsec:ablation}

\begin{figure*}[t]
    \centering
    \includegraphics[width=\textwidth]{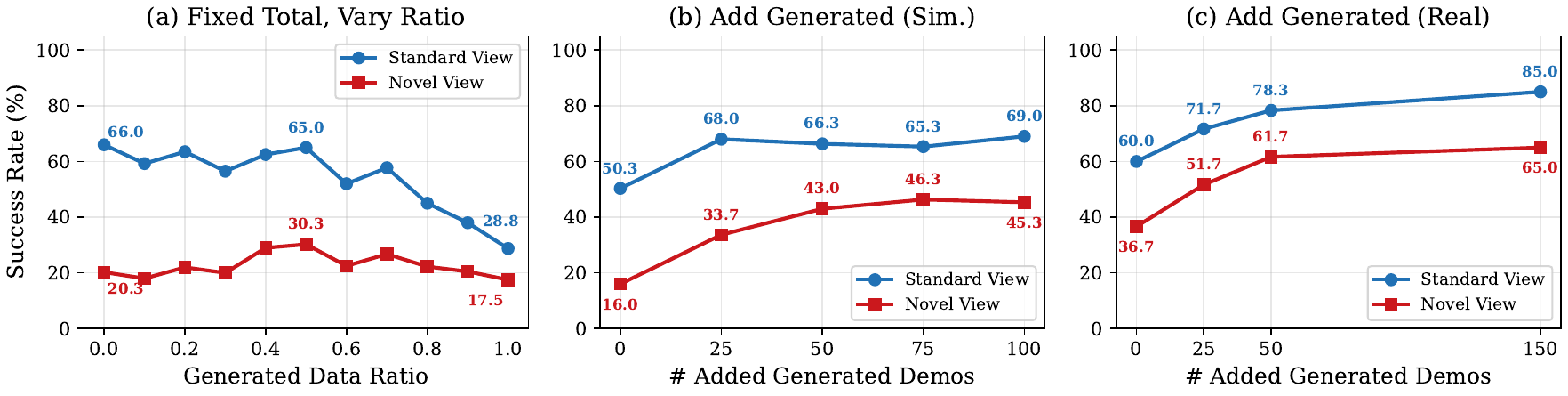}
    \caption{
        \textbf{Data scaling analysis.}
        (a)~Fixed total of 50 demonstrations with varying generated data ratio.
        (b)~Fixed 25 standard demonstrations with increasing generated data in simulation.
        (c)~Fixed 50 standard demonstrations with increasing generated data on the real robot.
    }
    \label{fig:data_scaling}
\end{figure*}

\begin{figure*}[t]
    \centering
    \includegraphics[width=\textwidth]{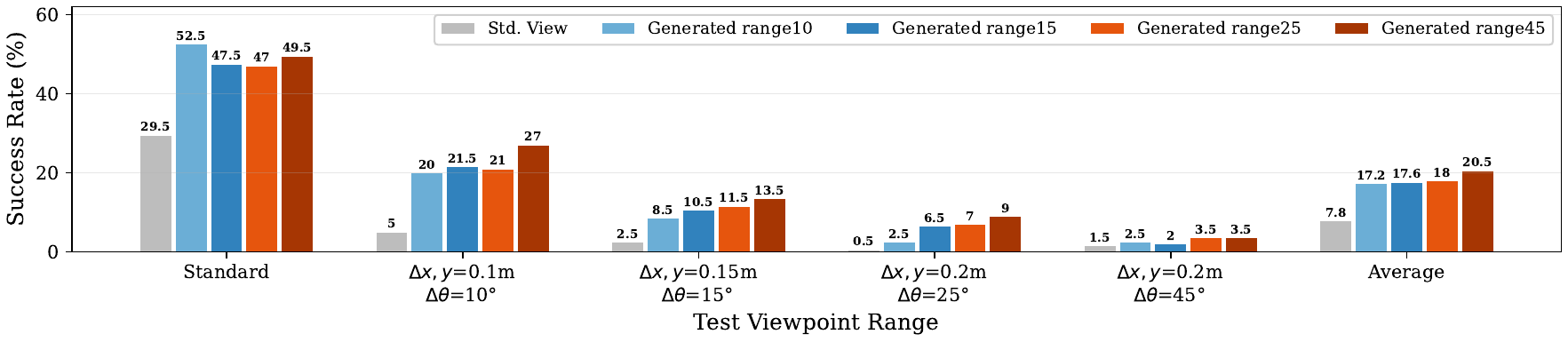}
    \caption{
        \textbf{Viewpoint generalization range analysis.}
        Each cluster represents a test viewpoint range; bars within each cluster correspond to different generation ranges used during training.
    }
    \label{fig:viewpoint_range}
\end{figure*}

\begin{table}[t]
  \centering
  \caption{\textbf{Ablation study: component analysis on three simulation tasks.}
  S/N: standard/novel egocentric viewpoint success rate (\%).
  FVD$\downarrow$: video generation quality.}
  \label{tab:ablation}
  \small
  \setlength{\tabcolsep}{2.5pt}
  \resizebox{\columnwidth}{!}{%
  \begin{tabular}{l cc cc cc cc c}
      \toprule
      & \multicolumn{2}{c}{\textit{Adjust}} 
      & \multicolumn{2}{c}{\textit{Lift}} 
      & \multicolumn{2}{c}{\textit{Place}} 
      & \multicolumn{2}{c}{\textbf{Avg.}}
      & \\
      \cmidrule(lr){2-3}
      \cmidrule(lr){4-5}
      \cmidrule(lr){6-7}
      \cmidrule(lr){8-9}
      \textbf{Variant}
      & S & N & S & N & S & N & S & N & \textbf{FVD}$\downarrow$ \\
      \midrule
      \rowcolor{gray!20}
      \textbf{EgoDemoGen}
      & \textbf{99} & \textbf{61} & \textbf{45} & 15 & \textbf{60} & \textbf{25} & \textbf{68.0} & \textbf{33.7} & \textbf{154.2} \\
      w/o Double Reproj.
      & 96 & 49 & 36 & 13 & 42 & 22 & 58.0 & 28.0 & 229.6 \\
      w/o Mask \& Inpaint
      & 99 & 53 & 42 & \textbf{16} & 57 & 21 & 66.0 & 30.0 & 205.8 \\
      w/o Traj. Transfer.
      & 96 & 40 & 37 & 9 & 51 & 8 & 61.3 & 19.0 & -- \\
      \bottomrule
  \end{tabular}%
  }
\end{table}

\begin{table}[t]
  \centering
  \caption{\textbf{Trajectory transfer validation:} replay success rate (\%) at novel egocentric viewpoints (50 trials per task).}
  \label{tab:trajectory_transfer}
  \small
  \begin{tabular}{l cccc}
      \toprule
      \textbf{Method} & \textit{Adjust} & \textit{Lift} & \textit{Place} & \textbf{Avg.} \\
      \midrule
      \rowcolor{gray!20}
      \textbf{EgoTrajTransfer} & \textbf{100} & \textbf{100} & \textbf{98} & \textbf{99.3} \\
      Source Action Trajectory & 52 & 8 & 8 & 22.7 \\
      \bottomrule
  \end{tabular}
\end{table}

We perform ablation studies to evaluate the contribution of each 
component of EgoDemoGen.

\Cref{tab:ablation} reports success rates and video quality (FVD) 
when removing individual components.
Removing trajectory transfer causes the largest novel-view drop, as 
the source actions are not suitable for the novel viewpoint, making the demonstrations ineffective for policy learning.
Removing double reprojection during training degrades both success 
rate and FVD, because the model has not learned to handle the reprojection artifacts.
Removing mask \& inpaint leaves residual robot artifacts in the scene video conditioning, which misleads the video generator and reduces novel-view performance.
All three components are necessary and contribute complementary benefits.
% End of Selection

\Cref{tab:trajectory_transfer} further validates trajectory 
transfer by replaying actions at novel egocentric viewpoints in 
open loop.
Transferred actions achieve near-perfect success whereas original 
actions largely fail, confirming that joint trajectories collected 
in the source egocentric frame are not directly reusable under base 
motion and that trajectory transfer is essential for constructing 
valid paired demonstrations.

\subsection{Data Scaling Analysis (\textbf{Q3})}
\label{subsec:data_scaling}

% \begin{figure*}[t]
%     \centering
%     \includegraphics[width=\textwidth]{newfigures/data_scaling.pdf}
%     \caption{
%         \textbf{Data scaling analysis.}
%         (a)~Fixed total of 50 demonstrations with varying generated data ratio.
%         (b)~Fixed 25 standard demonstrations with increasing generated data in simulation.
%         (c)~Fixed 50 standard demonstrations with increasing generated data on the real robot.
%     }
%     \label{fig:data_scaling}
% \end{figure*}

% \begin{figure*}[t]
%     \centering
%     \includegraphics[width=\textwidth]{newfigures/viewpoint_range.pdf}
%     \caption{
%         \textbf{Viewpoint generalization range analysis.}
%         Each cluster represents a test viewpoint range; bars within each cluster correspond to different generation ranges used during training.
%     }
%     \label{fig:viewpoint_range}
% \end{figure*}

\Cref{fig:data_scaling} examines how generated data quantity and ratio affect policy performance (averaged over tasks; per-task results in supplementary).
When fixing the total data budget and varying the generated ratio (\Cref{fig:data_scaling}a), novel-view performance peaks around 0.4--0.5 and declines as generated data dominates, suggesting that a balanced mixture is preferable to replacing all original demonstrations.
When adding generated demonstrations to a fixed set of original data (\Cref{fig:data_scaling}b,c), both standard- and novel-view success improve consistently in simulation and the real world, with the largest gains from the first batch of generated data and diminishing returns thereafter.

\subsection{Viewpoint Generalization Range (\textbf{Q4})}
\label{subsec:viewpoint_generalization}

% \begin{figure*}[t]
%     \centering
%     \includegraphics[width=\textwidth]{newfigures/viewpoint_range.pdf}
%     \caption{
%         \textbf{Viewpoint generalization range analysis} (averaged over two simulation tasks).
%         Each cluster represents a test viewpoint range; bars within each cluster correspond to different generation ranges used during training.
%         R$k$: $\Delta xy$\,m, $\Delta\theta$\,degrees as indicated.
%     }
%     \label{fig:viewpoint_range}
% \end{figure*}

\Cref{fig:viewpoint_range} examines how the generation viewpoint range affects generalization (averaged over two simulation tasks; per-task results in supplementary).
The average success rate increases monotonically from 7.8\% (standard view only) to 20.5\% as the generation range widens, with wider ranges especially beneficial at larger test shifts (e.g., at the 0.2\,m/25° test range, success improves from 0.5\% to 9.0\%); absolute success rates decrease with growing test range because larger shifts approach the robot's kinematic workspace limits, making tasks inherently harder.
Nevertheless, EgoDemoGen consistently improves over the standard-view baseline across all test ranges without degrading standard-view performance.

\subsection{Base-Camera Decoupled Generalization (\textbf{Q5})}
\label{subsec:decoupled_motion}

\begin{table}[t]
  \centering
  \caption{\textbf{Success rate (\%) under decoupled base and camera motion} on four simulation tasks.
  An additional random camera perturbation ($\Delta x', \Delta y' \in [-0.05, 0.05]$\,m, $\Delta \theta' \in [-10, 10]$°) is applied on top of the egocentric base motion.}
  \label{tab:decoupled}
  \small
  \begin{tabular}{l ccccc}
      \toprule
      \textbf{Method} & \textit{Adjust} & \textit{Lift} & \textit{Handover} & \textit{Place} & \textbf{Avg.} \\
      \midrule
      Std.\ View & 25 & 3 & 0 & 3 & 7.8 \\
      \rowcolor{gray!20}
      + \textbf{EgoDemoGen} & \textbf{51} & \textbf{10} & \textbf{13} & \textbf{19} & \textbf{23.3} \\
      \bottomrule
  \end{tabular}
\end{table}

In the preceding experiments, the egocentric camera is rigidly mounted on the robot base, so a base motion $(\Delta x, \Delta y, \Delta \theta)$ deterministically shifts both the action frame and the camera viewpoint.
Here we relax this assumption: on top of the base motion used in the main experiments ($\Delta x, \Delta y \in [-0.1, 0.1]$\,m, $\Delta \theta \in [-10, 10]$°), we apply an additional random camera perturbation ($\Delta x', \Delta y' \in [-0.05, 0.05]$\,m, $\Delta \theta' \in [-10, 10]$°) that is independent of the base motion.
This decouples the camera viewpoint change from the action-frame change, creating a more challenging setting where the two must be controlled separately.
EgoViewTransfer handles this naturally because it conditions camera viewpoint (via scene reprojection) and robot motion (via robot video rendering) through two independent inputs.
As shown in \Cref{tab:decoupled}, augmenting with EgoDemoGen-generated demonstrations under this decoupled setting still substantially improves novel-view success, confirming that the decoupled conditioning design generalizes beyond the rigidly-coupled assumption.

\section{Conclusion}
\label{sec:conclusion}

We present EgoDemoGen, a framework that generates paired observation and action demonstrations under novel egocentric viewpoints through two modules: EgoTrajTransfer for trajectory transfer via motion-skill segmentation, geometry-aware transformation, and inverse kinematics filtering, and EgoViewTransfer for observation video synthesis by fusing reprojected scene and rendered robot motion videos through a diffusion model trained with a self-supervised double reprojection strategy.
Experiments in simulation and real-world settings show that EgoDemoGen consistently improves policy success under both standard and novel egocentric viewpoints, outperforming geometry-based and video-generation-based methods.
Ablation studies and further analyses on data scaling, viewpoint generalization range, and base-camera decoupled settings confirm the scalability and flexibility of the framework.
In future work, we plan to integrate EgoDemoGen with world models for closed-loop demonstration generation, extend the paired generation paradigm to cross-embodiment transfer, and explore end-to-end training that jointly optimizes demonstration generation and downstream policy learning.

{
    \small
    \bibliographystyle{ieeenat_fullname}
    \bibliography{main}
}

\appendix
\clearpage
\setcounter{page}{1}
\section*{Supplementary Material of Egocentric Demonstration Generation for Viewpoint Generalization in Robotic Manipulation}

% \tableofcontents

%% ============================================================
%%  A. Method Details
%% ============================================================
\section{Method Details}
\label{sec:supp_method}

\subsection{Novel Viewpoint Parameters and Sampling}
\label{sec:supp_viewpoint}

An egocentric viewpoint change is parameterized by planar base motion $v = (\Delta x, \Delta y, \Delta \theta)$, where $\Delta x, \Delta y$ denote translational displacement in the world frame and $\Delta \theta$ denotes yaw rotation.
Because the egocentric camera is rigidly mounted on the robot head, which is fixed to the base, a base motion directly induces a corresponding change in the camera extrinsic matrix.

Concretely, given the camera-to-base extrinsic $T_{\text{cam}\leftarrow\text{base}} \!\in\! \text{SE}(3)$, the ego-motion parameters define a base displacement:
\begin{equation}
    \Delta T = 
    \begin{bmatrix}
        \cos\Delta\theta & -\sin\Delta\theta & 0 & \Delta x \\
        \sin\Delta\theta &  \cos\Delta\theta & 0 & \Delta y \\
        0 & 0 & 1 & 0 \\
        0 & 0 & 0 & 1
    \end{bmatrix}.
    \label{eq:delta_T}
\end{equation}
This transformation simultaneously changes: (1) the camera extrinsic, determining the novel viewpoint for observation synthesis, and (2) the robot base frame, requiring trajectory transfer for the action sequence.

During generation, for each source demonstration we sample one novel viewpoint by drawing $\Delta x, \Delta y$ uniformly from $[-r_{xy}, r_{xy}]$ and $\Delta \theta$ uniformly from $[-r_\theta, r_\theta]$, where $r_{xy}$ and $r_\theta$ are the generation range parameters.
In our main experiments, we use $r_{xy} = 0.1$\,m and $r_\theta = 10$\textdegree.
After sampling, the viewpoint undergoes feasibility filtering through the EgoTrajTransfer pipeline: viewpoints for which IK fails or produces excessively large joint jumps are discarded and new viewpoints are sampled until a valid viewpoint is found, ensuring that every generated demonstration is kinematically valid.

\subsection{EgoTrajTransfer: Detailed Algorithm}
\label{sec:supp_trajectory}

The main paper presents the high-level EgoTrajTransfer pipeline.
Here we provide the complete algorithm with dual-arm handling, coordinate system conversion, and feasibility filtering.
Algorithm~\ref{alg:trajectory_transfer_detail} summarizes the full five-step procedure.

The robot action is represented as a 14-dimensional joint vector $q_t \in \mathbb{R}^{14}$, partitioned as:
\begin{equation}
    q_t = [\underbrace{q_t^{L} \!\in\! \mathbb{R}^{6}}_{\text{left joints}},\; \underbrace{g_t^{L}}_{\text{grip}},\; \underbrace{q_t^{R} \!\in\! \mathbb{R}^{6}}_{\text{right joints}},\; \underbrace{g_t^{R}}_{\text{grip}}].
    \label{eq:joint_repr}
\end{equation}
All segmentation, transfer, and IK operations are performed independently for each arm, and gripper states are preserved without modification.

For \textbf{motion segments}, position is adapted by scaling and direction alignment (Eq.~\ref{eq:motion_pos_detail}), and orientation uses rotation-progress-based interpolation (Eq.~\ref{eq:motion_rot_detail}).
All formulas are annotated in Algorithm~\ref{alg:trajectory_transfer_detail}.

\begin{equation}
    p^{\text{new}}(t) = R_{\text{align}} \bigl(p^{\text{old}}(t) - p_{\text{start}}^{\text{old}}\bigr) \cdot s + p_{\text{start}}^{\text{new}},
    \label{eq:motion_pos_detail}
\end{equation}
\begin{equation}
    R^{\text{new}}(t) = \text{Slerp}(I, C; \alpha(t)) \cdot R_{\text{rel}}^{\text{old}}(t) \cdot R_{\text{start}}^{\text{new}},
    \label{eq:motion_rot_detail}
\end{equation}
where the auxiliary variables are:
\begin{gather}
    s = \frac{\|p_{\text{end}}^{\text{new}} \!-\! p_{\text{start}}^{\text{new}}\|}{\|p_{\text{end}}^{\text{old}} \!-\! p_{\text{start}}^{\text{old}}\|}, \;\;
    R_{\text{align}} = \text{Align}(\vec{d}^{\,\text{old}},\, \vec{d}^{\,\text{new}}), \notag \\
    C = R_{\delta}^{\text{new}} (R_{\delta}^{\text{old}})^{-1}, \;\;
    \alpha(t) = \frac{|\text{angle}(R_{\text{rel}}^{\text{old}}(t))|}{|\text{angle}(R_{\delta}^{\text{old}})|},
    \label{eq:motion_params}
\end{gather}
with $\vec{d}^{\,\text{old}} = p_{\text{end}}^{\text{old}} - p_{\text{start}}^{\text{old}}$ and $\vec{d}^{\,\text{new}} = p_{\text{end}}^{\text{new}} - p_{\text{start}}^{\text{new}}$.

For \textbf{skill segments}, the left arm transformation is directly $\Delta T_L$, while the right arm requires coordinate conversion through the camera-to-base extrinsics:
\begin{equation}
    \Delta T_R = T_{\text{c} \leftarrow \text{b}}^{R} (T_{\text{c} \leftarrow \text{b}}^{L})^{-1} \Delta T_L \, T_{\text{c} \leftarrow \text{b}}^{L} (T_{\text{c} \leftarrow \text{b}}^{R})^{-1},
    \label{eq:skill_right_convert}
\end{equation}
where $T_{\text{c} \leftarrow \text{b}}^{L}$ and $T_{\text{c} \leftarrow \text{b}}^{R}$ denote camera-to-base extrinsics for the left and right arms, respectively.
This conversion ensures the same base motion is correctly expressed in each arm's local coordinate frame.

\begin{algorithm*}[t]
\caption{EgoTrajTransfer: Detailed Dual-Arm Trajectory Transfer}
\label{alg:trajectory_transfer_detail}
\begin{algorithmic}[1]
\Require Source trajectory $Q = \{q_t\}_{t=1}^T$ ($q_t \in \mathbb{R}^{14}$: left arm 6 + gripper 1 + right arm 6 + gripper 1), base transformation $\Delta T$, camera-to-base extrinsics $T_{\text{c} \leftarrow \text{b}}^{L}, T_{\text{c} \leftarrow \text{b}}^{R}$, URDF $\mathcal{U}$, thresholds $\tau_{\text{IK}}, \tau_{\text{jump}}, \tau_{\text{close}}, w_{\text{sync}}$
\Ensure Transferred trajectory $\tilde{Q} = \{\tilde{q}_t\}_{t=1}^T$ or \texttt{INFEASIBLE}

\Statex \hspace{-1em}\textbf{// Step 1: Per-arm independent segmentation with boundary synchronization}
\State Split $q_t \rightarrow (q_t^L, g_t^L, q_t^R, g_t^R)$ for all $t$
\For{arm $a \in \{L, R\}$}
    \State Detect contact: $c_t^a = \mathbb{1}[g_t^a < \tau_{\text{close}}]$ \Comment{Closed gripper $\Rightarrow$ skill; open $\Rightarrow$ motion}
    \State Segment trajectory into alternating \texttt{motion} and \texttt{skill} segments based on $c_t^a$
\EndFor
\State Synchronize: merge transition boundaries within $w_{\text{sync}}$ frames between L/R arms \Comment{Avoid noise-induced misalignment}

\Statex \hspace{-1em}\textbf{// Step 2: Compute arm-specific rigid transformations}
\State $\Delta T_L \gets \Delta T$ \Comment{Left arm: direct application in left base frame}
\State $\Delta T_R \gets T_{\text{c} \leftarrow \text{b}}^{R} (T_{\text{c} \leftarrow \text{b}}^{L})^{-1} \Delta T_L \, T_{\text{c} \leftarrow \text{b}}^{L} (T_{\text{c} \leftarrow \text{b}}^{R})^{-1}$ \Comment{Right arm: convert via camera extrinsics}

\Statex \hspace{-1em}\textbf{// Step 3: Transfer each arm's segments independently}
\For{arm $a \in \{L, R\}$ with transformation $\Delta T_a$}
    \State Compute FK: $\{T_e^a(t)\} \gets \text{FK}_a(\{q_t^a\})$ \Comment{End-effector pose sequence}
    \For{each segment $s$ of arm $a$}
        \If{$s$ is \texttt{motion} segment} \Comment{Free-space movement: scale + align + interpolate}
            \State Compute new endpoint: $T_{\text{end}}^{\text{new}} = \Delta T_a \cdot T_{\text{end}}^{\text{old}}$; new start from preceding segment
            \State $s \gets \|p_{\text{end}}^{\text{new}} - p_{\text{start}}^{\text{new}}\| / \|p_{\text{end}}^{\text{old}} - p_{\text{start}}^{\text{old}}\|$ \Comment{Path length scale}
            \State $R_{\text{align}} \gets \text{Align}(p_{\text{end}}^{\text{old}} - p_{\text{start}}^{\text{old}},\; p_{\text{end}}^{\text{new}} - p_{\text{start}}^{\text{new}})$ \Comment{Direction alignment rotation}
            \State $p^{\text{new}}(t) = R_{\text{align}} (p^{\text{old}}(t) - p_{\text{start}}^{\text{old}}) \cdot s + p_{\text{start}}^{\text{new}}$ \Comment{Position: scale \& align}
            \State $\alpha(t) \gets |\text{angle}(R_{\text{rel}}^{\text{old}}(t))| \,/\, |\text{angle}(R_{\delta}^{\text{old}})|$ \Comment{Rotation progress ratio}
            \State $C \gets R_{\delta}^{\text{new}} (R_{\delta}^{\text{old}})^{-1}$ \Comment{Correction: new vs old total rotation}
            \State $R^{\text{new}}(t) = \text{Slerp}(I, C;\, \alpha(t)) \cdot R_{\text{rel}}^{\text{old}}(t) \cdot R_{\text{start}}^{\text{new}}$ \Comment{Orientation: progress-based interpolation}
        \ElsIf{$s$ is \texttt{skill} segment} \Comment{Contact-rich: rigid transform preserves object-relative motion}
            \State $T_e^{a,\text{new}}(t) = \Delta T_a \cdot T_e^{a,\text{old}}(t)$ for all $t$ in segment
        \EndIf
    \EndFor
\EndFor

\Statex \hspace{-1em}\textbf{// Step 4: Dual-arm IK solving and feasibility filtering}
\For{arm $a \in \{L, R\}$}
    \State $\{\tilde{q}_t^a\} \gets \text{CuRobo\_BatchIK}(\{T_e^{a,\text{new}}(t)\},\; \text{seed} = \{q_t^a\})$ \Comment{Original joints as seeds for config continuity}
    \State Linearly interpolate failed frames from nearest successful neighbors \Comment{Ensure trajectory continuity}
\EndFor
\If{IK success rate $< \tau_{\text{IK}}$ for either arm} \Comment{Task exceeds workspace after base motion}
    \State \Return \texttt{INFEASIBLE}
\EndIf
\If{$\max_t \|\tilde{q}_{t+1}^a - \tilde{q}_t^a\|_\infty > \tau_{\text{jump}}$ for either arm} \Comment{IK solver discontinuity}
    \State \Return \texttt{INFEASIBLE}
\EndIf

\Statex \hspace{-1em}\textbf{// Step 5: Reassemble dual-arm trajectory}
\State $\tilde{q}_t \gets [\tilde{q}_t^L,\; g_t^L,\; \tilde{q}_t^R,\; g_t^R]$ for all $t$ \Comment{Concatenate with preserved original grippers}
\State \Return $\tilde{Q} = \{\tilde{q}_t\}_{t=1}^T$
\end{algorithmic}
\end{algorithm*}

\subsection{Novel Viewpoint Reprojection}
\label{sec:supp_reprojection}

We reproject original RGB-D frames to the novel viewpoint via a backproject-transform-project pipeline, as detailed in Algorithm~\ref{alg:reproj}.
This process introduces holes (from newly revealed occluded regions) and stretching artifacts (near depth discontinuities), which are handled by the subsequent mask-and-inpaint step and the EgoViewTransfer model.
For double reprojection training, this pipeline is applied twice (source $\rightarrow$ novel $\rightarrow$ source) to amplify artifacts that the model learns to repair.

\subsection{Robot Motion Video and Mask Rendering}
\label{sec:supp_rendering}

We render a robot motion video $V_R$ and binary robot mask $\mathcal{M}$ from joint trajectories using off-screen rasterization with the robot's URDF model, as detailed in Algorithm~\ref{alg:render}.
The robot mask of the original trajectory under the novel viewpoint removes the robot from the reprojected scene video; the rendered robot motion video of the transferred trajectory provides explicit motion conditioning for the video generator.

\begin{algorithm*}[t]
\caption{RGB-D Novel-View Reprojection}
\label{alg:reproj}
\begin{algorithmic}[1]
\Require RGB-D sequence $\{I_t, D_t\}_{t=0}^{T-1}$ from source viewpoint, camera intrinsics $K \in \mathbb{R}^{3 \times 3}$, viewpoint transformation $T_{0 \rightarrow 1} \in \text{SE}(3)$, robot mask $\{M_t\}$ (optional), depth scale $s$ (mm $\rightarrow$ m), max depth $d_{\max}$
\Ensure Novel-view RGB and depth $\{\hat{I}_t, \hat{D}_t\}_{t=0}^{T-1}$
\For{$t = 0$ to $T{-}1$}
    \State $D_t^{\text{rgb}} \gets \text{AlignDepthToRGB}(D_t, I_t, s)$ \Comment{Resample depth to match RGB resolution and coordinate}
    \State $(I_t^{\star}, D_t^{\star}) \gets \text{ApplyMask}(I_t, D_t^{\text{rgb}}, M_t)$ \Comment{Optional: mask robot region with spatial dilation}
    \State $(\mathbf{P}_t, \mathbf{C}_t) \gets \text{Backproject}(I_t^{\star}, D_t^{\star}, K, s, d_{\max})$ \Comment{$\mathbf{P}_t \in \mathbb{R}^{N \times 3}$: 3D points; $\mathbf{C}_t$: per-point RGB}
    \State $\mathbf{P}_t^{(1)} \gets T_{0 \rightarrow 1} \, \mathbf{P}_t$ \Comment{Rigid transform point cloud to target camera frame}
    \State $(\hat{I}_t, \hat{D}_t) \gets \text{ZBufferSplat}(\mathbf{P}_t^{(1)}, \mathbf{C}_t, K)$ \Comment{Bilinear splatting with per-pixel min-depth for occlusion}
\EndFor
\Statex {\small$\triangleright$ Frames are processed independently. A second pass via $T_{1 \rightarrow 0}$ produces double-reprojection training data.}
\State \Return $\{\hat{I}_t, \hat{D}_t\}_{t=0}^{T-1}$
\end{algorithmic}
\end{algorithm*}

\begin{algorithm*}[t]
\caption{Robot Motion Video and Mask Rendering}
\label{alg:render}
\begin{algorithmic}[1]
\Require URDF model $\mathcal{U}$ with link meshes and materials, joint trajectory $\{q_t\}_{t=0}^{T-1}$, camera intrinsics $K$, camera extrinsic $T_{\text{cam} \leftarrow \text{base}}$, image size $(H, W)$
\Ensure Robot motion video $V_R$ (RGB) and binary robot mask $\mathcal{M}$
\State Load robot from $\mathcal{U}$; cache per-link collision/visual meshes and materials; instantiate FK solver
\State Configure off-screen renderer: set camera $(K, T_{\text{cam} \leftarrow \text{base}})$, image size $(H, W)$, lighting, and z-buffer
\For{$t = 0$ to $T{-}1$}
    \State $\{L_i(t)\}_{i=1}^{N_L} \gets \text{FK}(q_t)$ \Comment{Forward kinematics: per-link 6-DoF poses in the base frame}
    \State $L_i^{\text{cam}}(t) \gets T_{\text{cam} \leftarrow \text{base}} \, L_i(t)$ for all links $i$ \Comment{Transform to camera frame}
    \State Rasterize robot geometry only (transparent background) with z-buffer and anti-aliasing $\rightarrow$ $\text{RGB}_t$, $\alpha_t$
    \State $\text{Mask}_t \gets \mathbb{1}[\alpha_t > 0]$ \Comment{Binary mask from alpha channel}
\EndFor
\State $V_R \gets \text{Encode}(\{\text{RGB}_t\})$; \quad $\mathcal{M} \gets \text{Stack}(\{\text{Mask}_t\})$
\Statex {\small$\triangleright$ Used in two roles: (1) mask from original trajectory at novel view removes robot from scene video; (2) robot motion video from transferred trajectory at novel view provides motion conditioning.}
\State \Return $V_R$, $\mathcal{M}$
\end{algorithmic}
\end{algorithm*}

\subsection{EgoViewTransfer: Model Architecture}
\label{sec:supp_architecture}

EgoViewTransfer is built on CogVideoX-5B-I2V~\cite{yang2024cogvideox}, a DiT-based image-to-video diffusion model with a 3D VAE.
The original model accepts a noise latent and one conditional image latent (each 16 channels after VAE encoding), concatenated along the channel dimension before the DiT's patch embedding layer (32 channels total).

To support dual-video conditioning (scene video $V_S$ and robot motion video $V_R$), we expand the patch embedding input from 32 to 48 channels.
The three 16-channel latents are concatenated as:
\begin{equation}
    z_{\text{in}} = [z_{\text{noise}},\; z_{V_S},\; z_{V_R}] \in \mathbb{R}^{T' \!\times\! H' \!\times\! W' \!\times\! 48},
    \label{eq:concat}
\end{equation}
where $z_{V_S}\!=\!\text{Enc}(V_S)$ and $z_{V_R}\!=\!\text{Enc}(V_R)$ are VAE-encoded scene and robot motion video latents, and $T', H', W'$ are the latent dimensions.
The additional 16 channels are zero-initialized so finetuning starts from the pretrained I2V behavior.
Only the DiT is finetuned; the 3D VAE remains frozen.
Training and inference hyperparameters are provided in the implementation details (Sec.~\ref{sec:supp_sim} and~\ref{sec:supp_real}).

\subsection{Policy Training with Generated Data}
\label{sec:supp_policy_training}

After generating novel egocentric demonstrations $\tilde{\mathcal{D}}$ via EgoDemoGen, we combine them with the original demonstrations $\mathcal{D}$ to form the augmented training set $\mathcal{D}_{\text{aug}} = \mathcal{D} \cup \tilde{\mathcal{D}}$.
The downstream policy is trained on $\mathcal{D}_{\text{aug}}$ via standard behavior cloning without any modification to the policy architecture.
Generated demonstrations are treated identically to original ones.
In main experiments, we use $M = N$ (1:1 ratio); the data scaling analysis explores other ratios.

%% ============================================================
%%  B. Simulation Implementation Details
%% ============================================================
% \clearpage
\section{Simulation Implementation Details}
\label{sec:supp_sim}

\subsection{Robot Configuration}
\label{sec:supp_sim_robot}

\begin{figure}[t]
    \centering
    \includegraphics[width=\columnwidth]{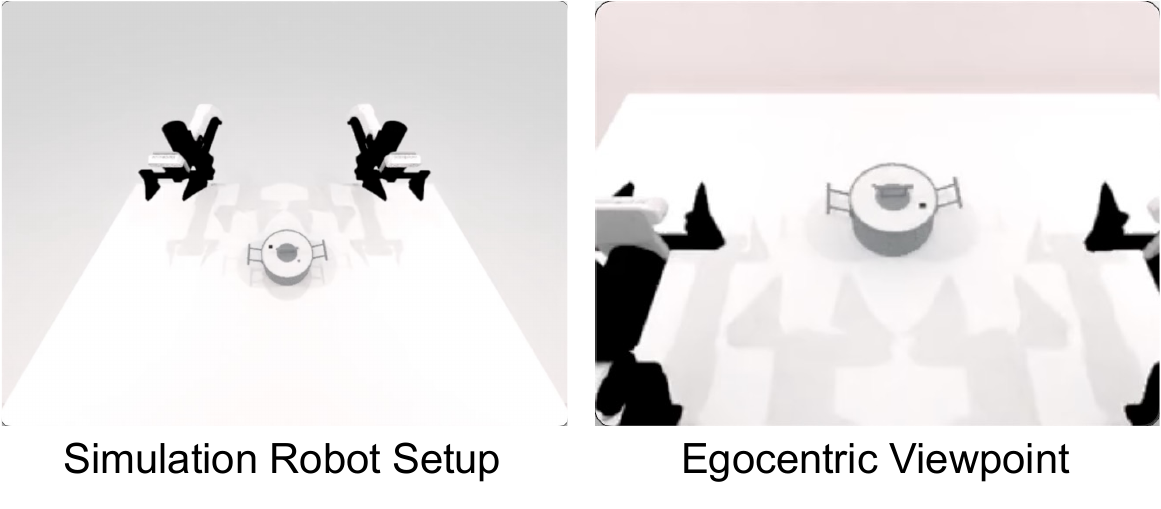}
    \caption{
        Simulation robot configuration and egocentric viewpoint.
        \textbf{Left}: The RoboTwin 2.0 environment with dual ARX-X5 manipulators.
        \textbf{Right}: Illustration of the head-mounted egocentric viewpoint.
    }
    \label{fig:sim_robot}
\end{figure}

As shown in \Cref{fig:sim_robot}, we use the RoboTwin 2.0~\cite{chen2025robotwin} simulation environment with dual ARX-X5 manipulators.
The platform is equipped with one head-mounted egocentric RGB-D camera (resolution $240 \times 320$, vertical FOV $53$\textdegree), which serves as the observation input.

\subsection{EgoViewTransfer Training}
\label{sec:supp_sim_egoview}

The simulation EgoViewTransfer model is trained on 500 episodes from 50 RoboTwin 2.0 tasks (10 episodes per task), all collected under the standard egocentric viewpoint.
Double reprojection training pairs are constructed by sampling viewpoints within $\Delta x, \Delta y \in [-0.2, 0.2]$\,m and $\Delta \theta \in [-45, 45]$\textdegree.
Training uses AdamW optimizer (lr $= 2 \times 10^{-5}$), batch size 2 per GPU, gradient accumulation over 8 steps, for 100 epochs on 4$\times$H20 GPUs.
Input resolution is $240 \times 320$ with 49 frames per clip.
During inference, long videos are processed as 49-frame segments with 25 DPM-Solver denoising steps.

\subsection{Policy Training}
\label{sec:supp_sim_policy}

We use ACT~\cite{zhao2023learning} with default hyperparameters: batch size 16, learning rate $1 \times 10^{-5}$, action chunk size 50, input resolution $480 \times 640$.
Training runs for 40k steps on 1$\times$RTX 4090 GPU.
During evaluation, temporal aggregation is used, and each method is evaluated over 100 trials per task.

\subsection{Task Rollout Examples}
\label{sec:supp_sim_rollout}

\begin{figure}[t]
    \centering
    \includegraphics[width=\columnwidth]{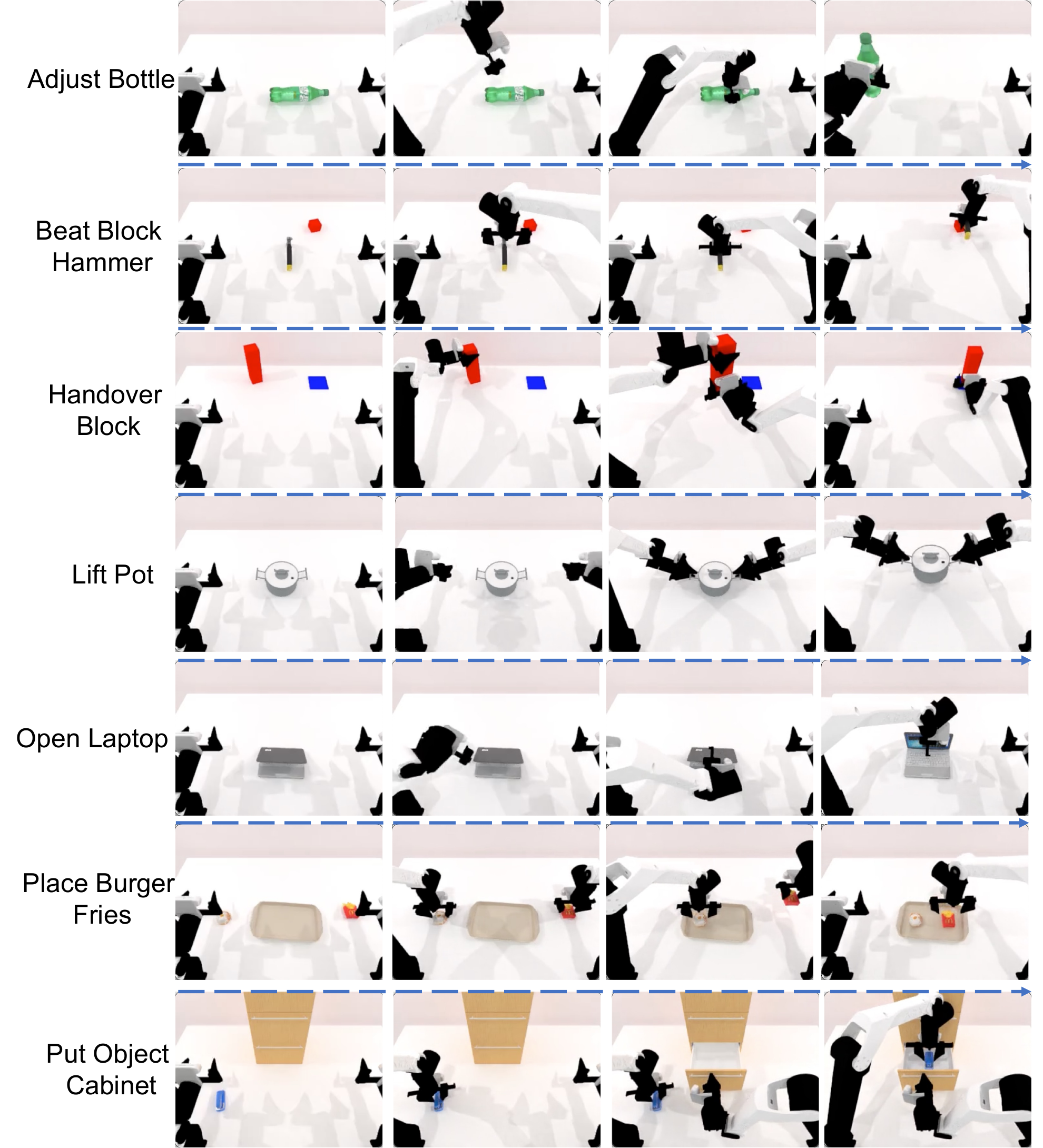}
    \caption{
        Task rollout examples in simulation.
    }
    \label{fig:sim_rollout}
\end{figure}

\Cref{fig:sim_rollout} shows simulation task rollout examples.

%% ============================================================
%%  C. Real-World Implementation Details
%% ============================================================
\section{Real-World Implementation Details}
\label{sec:supp_real}

\subsection{Robot Configuration}
\label{sec:supp_real_robot}

\begin{figure}[t]
    \centering
    \includegraphics[width=\columnwidth]{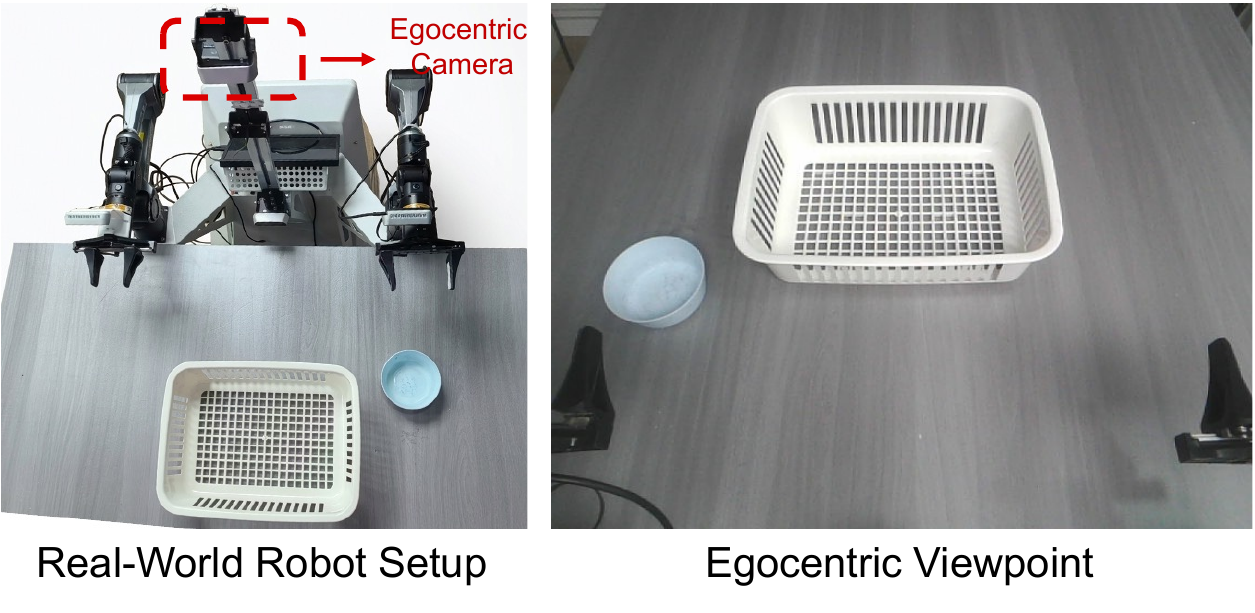}
    \caption{
        Real-world robot configuration and egocentric viewpoint.
        \textbf{Left}: The Mobile ALOHA platform with dual arms.
        \textbf{Right}: Illustration of the head-mounted egocentric viewpoint.
    }
    \label{fig:real_robot}
\end{figure}

As shown in \Cref{fig:real_robot}, we use the Mobile ALOHA platform with dual arms, equipped with one head-mounted egocentric RGB-D camera (Intel RealSense D435i, RGB resolution $480 \times 640$, depth resolution $400 \times 640$, vertical FOV $53$\textdegree), which serves as the observation input.
The mismatched RGB and depth resolutions cause additional alignment artifacts in the reprojected scene video compared to the simulation setting, yet EgoViewTransfer still produces high-quality novel-viewpoint observations, demonstrating the robustness of the double reprojection self-supervised training strategy.

\subsection{EgoViewTransfer Training}
\label{sec:supp_real_egoview}

The real-world EgoViewTransfer model is trained on approximately 600 teleoperated episodes spanning diverse manipulation scenarios (tabletop pick-and-place, microwave open/close, sink cleaning, cloth folding), all collected under the standard egocentric viewpoint.
For episodes missing depth, scale-consistent depth maps are generated using MoGe~\cite{wang2025moge}.
Double reprojection training uses $\Delta x, \Delta y \in [-0.2, 0.2]$\,m and $\Delta \theta \in [-30, 30]$\textdegree.
Training uses AdamW optimizer (lr $= 2 \times 10^{-5}$), batch size 1 per GPU, gradient accumulation over 8 steps, for 120 epochs on 4$\times$H20 GPUs.
Input resolution is $480 \times 640$ with 49 frames per clip.
Inference settings are the same as simulation (49-frame segments, 25 DPM-Solver steps).

\subsection{Policy Training}
\label{sec:supp_real_policy}

We use $\pi_0$~\cite{black2024pi0visionlanguageactionflowmodel} with default settings: batch size 32, input resolution $480 \times 640$, learning rate $2.5 \times 10^{-5}$, action chunk size 50, training for 20k steps on 4$\times$H20 GPUs.
Each method is evaluated over 20 trials per task for the standard viewpoint and four novel viewpoints.

\subsection{Task Rollout Examples}
\label{sec:supp_real_rollout}

\begin{figure}[t]
    \centering
    \includegraphics[width=\columnwidth]{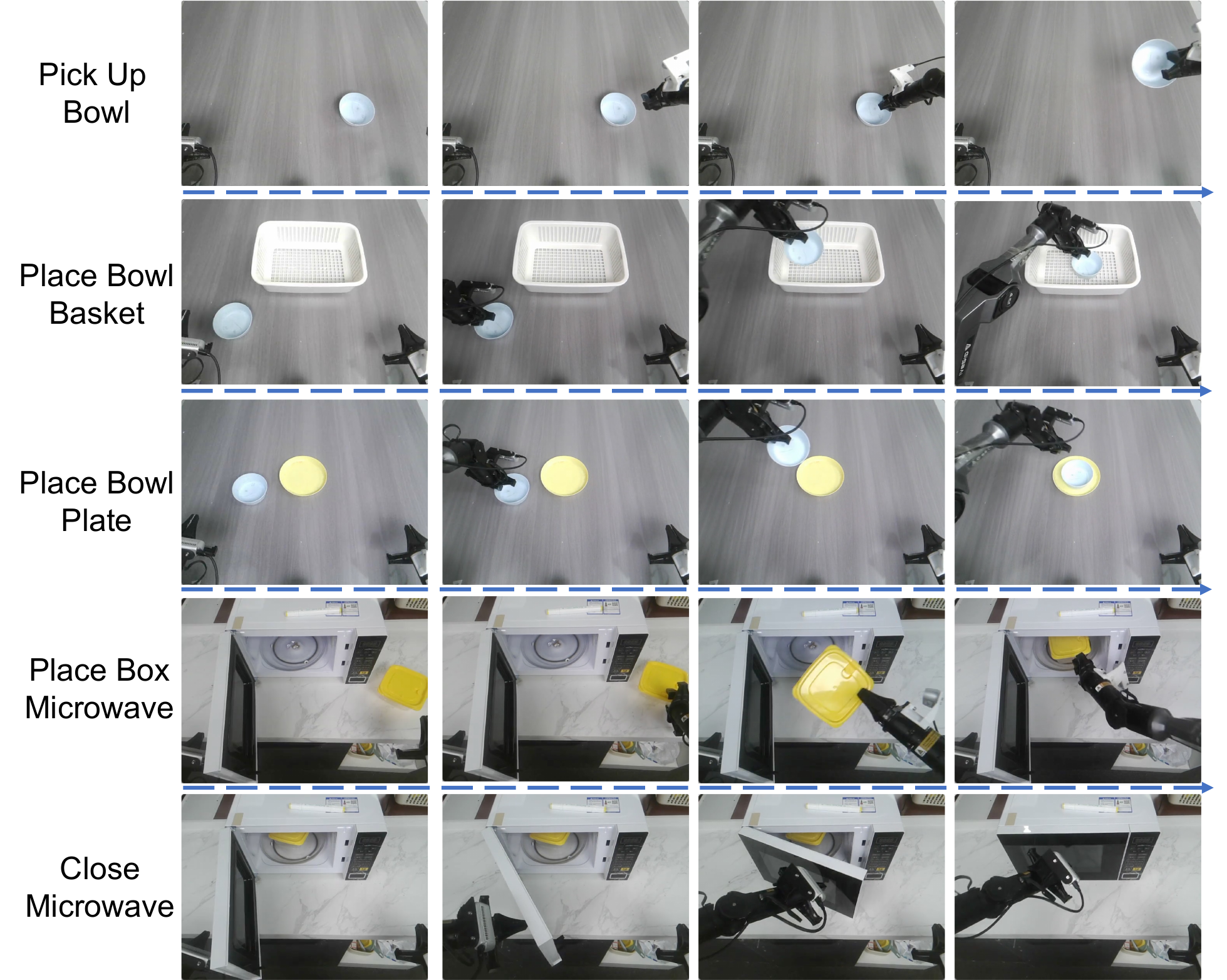}
    \caption{
        Task rollout examples in the real world.
    }
    \label{fig:real_rollout}
\end{figure}

\Cref{fig:real_rollout} shows real-world task rollout examples.

%% ============================================================
%%  D. Additional Experimental Results
%% ============================================================
\section{Additional Experimental Results}
\label{sec:supp_results}

\subsection{Detailed Data Scaling Results}
\label{sec:supp_data_scaling}

The data scaling analysis in the main paper reports task-averaged results.
Here we provide per-task breakdowns for both the fixed-total varying-ratio experiment and the fixed-original adding-generated experiment.

\begin{figure}[t]
    \centering
    \includegraphics[width=\columnwidth]{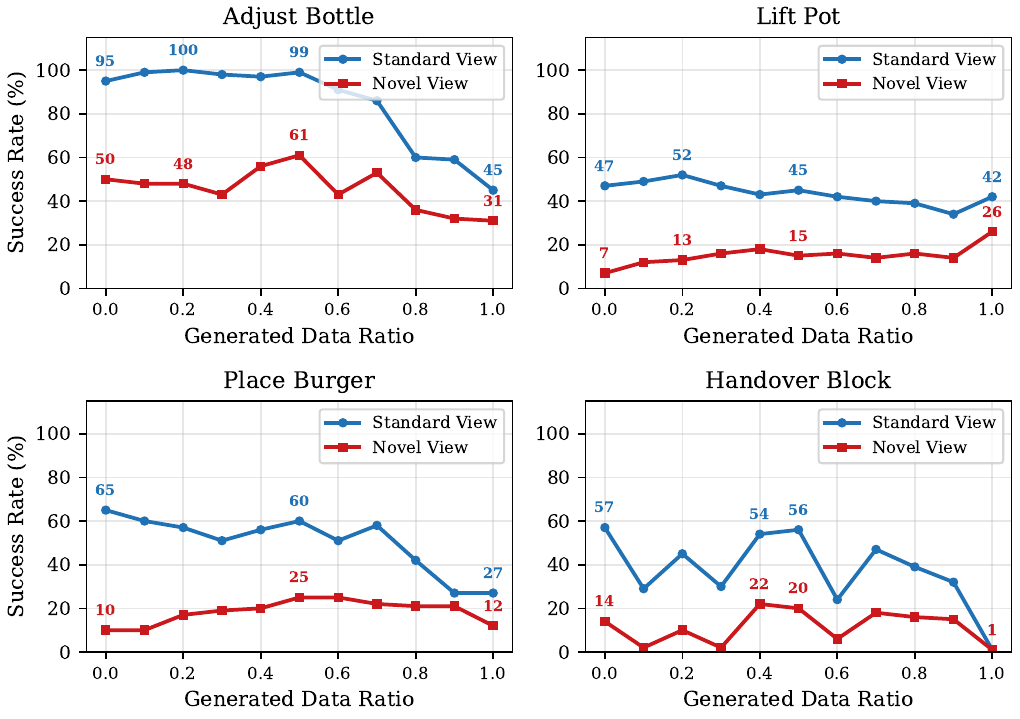}
    \caption{
        Per-task data scaling results with fixed total demonstrations and varying generated data ratio (simulation).
    }
    \label{fig:scaling_pertask_ratio}
\end{figure}

\noindent\textbf{Fixed total, varying generated ratio.}
\Cref{fig:scaling_pertask_ratio} shows per-task success rates as the generated data ratio increases from 0 to 1.0 (total demonstrations fixed at 50).
For most tasks, both standard-view and novel-view performance peak at moderate ratios (0.3--0.5) and degrade when the generated proportion is too high.
The trend is especially pronounced for Adjust Bottle, where the standard-view rate drops from 95\% to 45\% at ratio 1.0, indicating that entirely replacing original demonstrations with generated ones hurts policy learning.
Handover Block exhibits the highest variance, reflecting the sensitivity of bimanual coordination tasks to data composition.

\begin{figure}[t]
    \centering
    \includegraphics[width=\columnwidth]{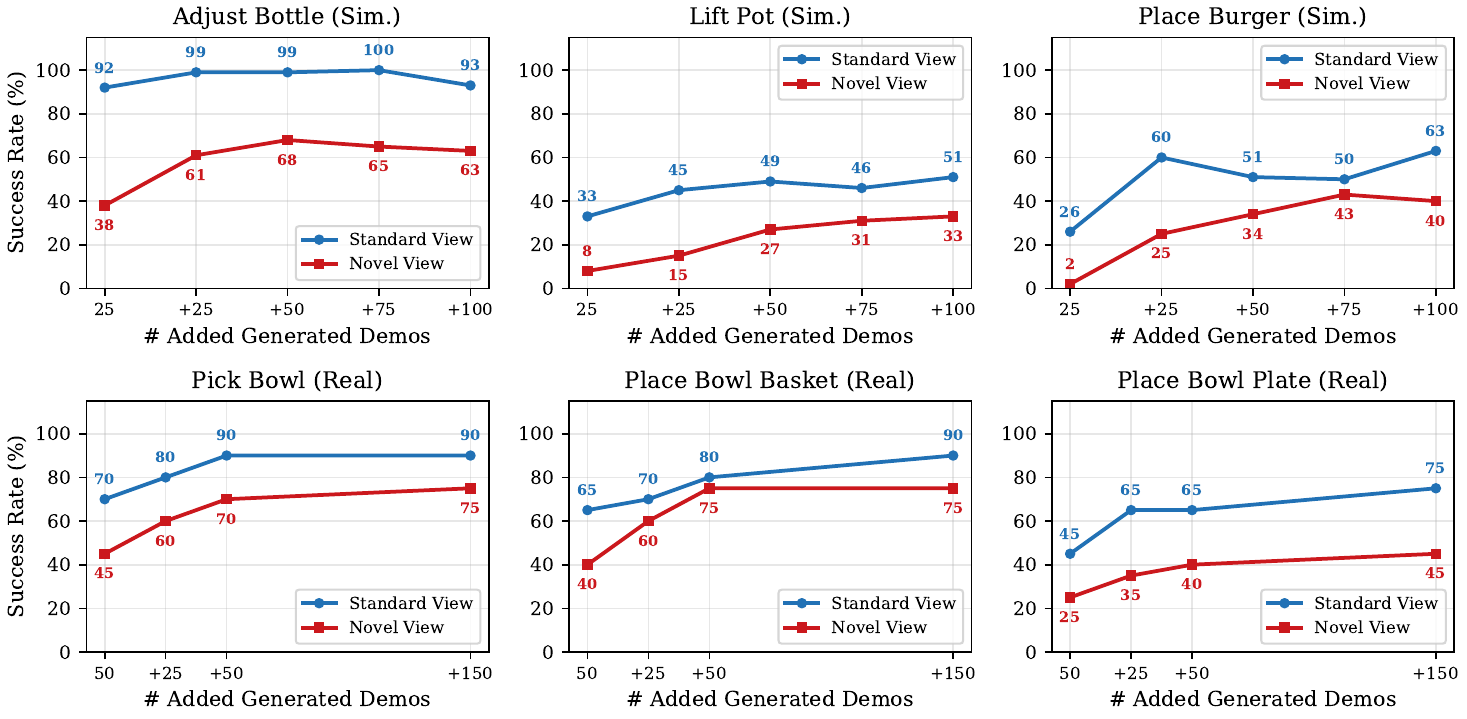}
    \caption{
        Per-task data scaling results with fixed original demonstrations and incrementally added generated demonstrations. Top row: simulation (3 tasks). Bottom row: real world (3 tasks).
    }
    \label{fig:scaling_pertask_add}
\end{figure}

\noindent\textbf{Fixed original, adding generated demonstrations.}
\Cref{fig:scaling_pertask_add} shows per-task results when progressively adding generated demonstrations on top of the original data.
In simulation, all three tasks benefit from adding generated data, with novel-view success rates improving consistently (e.g., Place Burger: 2\% $\rightarrow$ 40\%).
In the real world, all three tasks show steady gains in both standard-view and novel-view performance as more generated demonstrations are added, confirming the effectiveness of EgoDemoGen across different manipulation scenarios.

\subsection{Detailed Viewpoint Generalization Results}
\label{sec:supp_viewpoint_range}

The viewpoint generalization analysis in the main paper reports results averaged over two simulation tasks.
Here we provide per-task breakdowns of policy success rates, trajectory replay success rates, and video generation quality across different generation ranges.

\noindent\textbf{Per-task policy success rates.}
\Cref{tab:vp_lift_pot,tab:vp_place_burger} report per-task success rates under different combinations of generation range (rows) and test viewpoint range (columns).
For Lift Pot (\Cref{tab:vp_lift_pot}), widening the generation range consistently improves performance under larger test shifts (e.g., the success rate at the 0.2\,m/25° test range improves from 1\% to 12\% with range45 generation).
Standard-view performance also improves from 33\% to 44--46\% with any generated data, indicating that viewpoint-augmented demonstrations benefit in-distribution performance as well.
For Place Burger (\Cref{tab:vp_place_burger}), a similar trend holds: range45 generation achieves the best performance under the widest test shifts (32\% at 0.1\,m/10° vs.\ 2\% baseline), though all generation ranges substantially improve over the standard-view-only baseline.

\begin{table}[t]
  \centering
  \caption{Per-task viewpoint generalization: \textbf{Lift Pot} success rate (\%). Rows: generation range used during training. Columns: test viewpoint range.}
  \label{tab:vp_lift_pot}
  \small
  \setlength{\tabcolsep}{3.5pt}
  \begin{tabular}{l ccccc}
      \toprule
      \multirow{2}{*}{\textbf{Gen.\ Range}} & \multicolumn{5}{c}{\textbf{Test Viewpoint Range}} \\
      \cmidrule(lr){2-6}
       & Std. & \makecell{0.1m\\10°} & \makecell{0.15m\\15°} & \makecell{0.2m\\25°} & \makecell{0.2m\\45°} \\
      \midrule
      None (Std.\ only) & 33 & 8 & 4 & 1 & 2 \\
      range10 & 45 & 15 & 9 & 4 & 4 \\
      range15 & \textbf{46} & 15 & 12 & 5 & 4 \\
      range25 & 44 & 17 & 14 & 8 & \textbf{6} \\
      range45 & 44 & \textbf{22} & \textbf{15} & \textbf{12} & 5 \\
      \bottomrule
  \end{tabular}
\end{table}

\begin{table}[t]
  \centering
  \caption{Per-task viewpoint generalization: \textbf{Place Burger} success rate (\%). Rows: generation range used during training. Columns: test viewpoint range.}
  \label{tab:vp_place_burger}
  \small
  \setlength{\tabcolsep}{3.5pt}
  \begin{tabular}{l ccccc}
      \toprule
      \multirow{2}{*}{\textbf{Gen.\ Range}} & \multicolumn{5}{c}{\textbf{Test Viewpoint Range}} \\
      \cmidrule(lr){2-6}
       & Std. & \makecell{0.1m\\10°} & \makecell{0.15m\\15°} & \makecell{0.2m\\25°} & \makecell{0.2m\\45°} \\
      \midrule
      None (Std.\ only) & 26 & 2 & 1 & 0 & 1 \\
      range10 & \textbf{60} & 25 & 8 & 1 & 1 \\
      range15 & 49 & 28 & 9 & \textbf{8} & 0 \\
      range25 & 50 & 25 & 9 & 6 & 1 \\
      range45 & 55 & \textbf{32} & \textbf{12} & 6 & \textbf{2} \\
      \bottomrule
  \end{tabular}
\end{table}

\begin{figure}[t]
    \centering
    \includegraphics[width=\columnwidth]{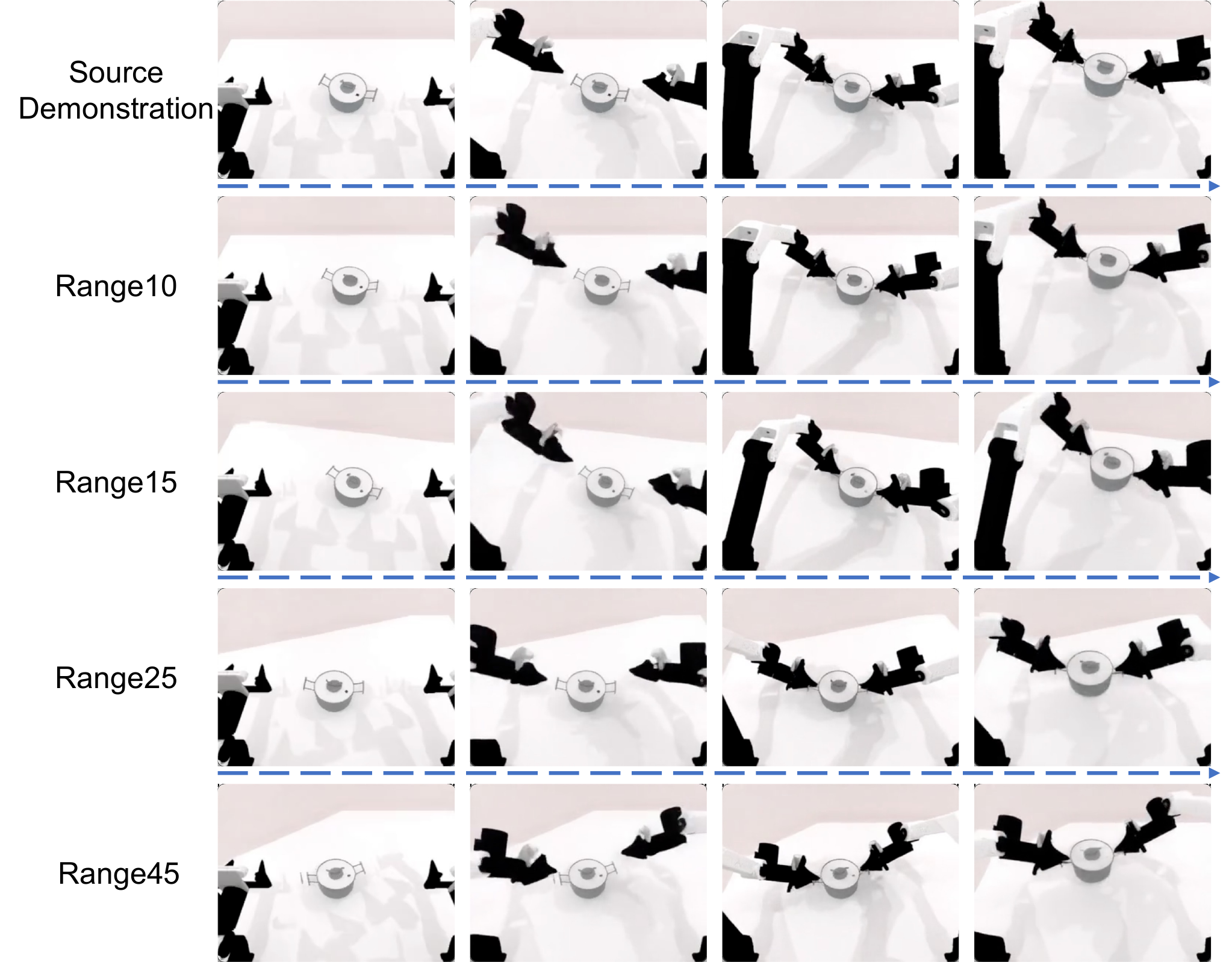}
    \caption{
        EgoViewTransfer generation visualization across different viewpoint ranges.
        From top to bottom: source demonstration (standard viewpoint), and generated videos at range10, range15, range25, and range45.
        The task-relevant central region remains well-synthesized across all ranges, while boundary artifacts increase with larger viewpoint shifts.
    }
    \label{fig:range_vis}
\end{figure}

\noindent\textbf{Trajectory replay success rate and video generation quality.}
\Cref{tab:vp_replay_quality} reports the EgoTrajTransfer replay success rate and EgoViewTransfer video generation quality across generation ranges.
The replay success rate remains near 100\% (100\% for Lift Pot, 98\% for Place Burger) regardless of the generation range, because infeasible viewpoints are already filtered by the IK feasibility check during trajectory transfer.
Video generation quality shows a slight degradation as the generation range increases (PSNR from 25.37 to 24.16, FVD from 154.2 to 185.4).
This is primarily because larger viewpoint shifts introduce wider boundary regions that require inpainting in the reprojected scene video; these peripheral areas are more difficult for EgoViewTransfer to reconstruct faithfully.
However, as visualized in \Cref{fig:range_vis}, the task-relevant central regions (robot arms, manipulated objects, and workspace) are consistently well-synthesized across all generation ranges, which explains why the downstream policy success rates still improve monotonically despite the slight quality drop in boundary areas.

\begin{table}[t]
  \centering
  \caption{Trajectory replay success rate (\%) and video generation quality across generation ranges (averaged over Lift Pot and Place Burger).}
  \label{tab:vp_replay_quality}
  \small
  \setlength{\tabcolsep}{4pt}
  \begin{tabular}{l cc cccc}
      \toprule
      \multirow{2}{*}{\textbf{Gen.\ Range}} & \multicolumn{2}{c}{\textbf{Replay (\%)}} & \multicolumn{4}{c}{\textbf{Video Quality}} \\
      \cmidrule(lr){2-3} \cmidrule(lr){4-7}
       & Lift & Burger & PSNR$\uparrow$ & SSIM$\uparrow$ & LPIPS$\downarrow$ & FVD$\downarrow$ \\
      \midrule
      range10 & 100 & 98 & 25.37 & 0.881 & 0.080 & 154.2 \\
      range15 & 100 & 98 & 24.89 & 0.875 & 0.088 & 169.8 \\
      range25 & 100 & 98 & 24.55 & 0.871 & 0.093 & 179.5 \\
      range45 & 100 & 98 & 24.16 & 0.866 & 0.102 & 185.4 \\
      \bottomrule
  \end{tabular}
\end{table}

\subsection{Results with More Initial Standard-Viewpoint Demonstrations}
\label{sec:supp_more_demos}

We investigate whether the benefit of EgoDemoGen persists when more initial standard-viewpoint demonstrations are available.
\Cref{tab:more_demos} reports the average success rate over four simulation tasks (Adjust Bottle, Lift Pot, Place Burger, and Handover Block) when training ACT policies with 25, 50, 200, and 1000 initial demonstrations, with and without EgoDemoGen augmentation.

As the number of initial demonstrations increases, standard-view performance improves steadily (38.0\% $\rightarrow$ 75.8\%), yet novel egocentric viewpoint performance plateaus around 31\% because additional single-viewpoint data does not introduce egocentric viewpoint diversity.
Augmenting with EgoDemoGen yields consistent novel-view gains at all data scales: +18.0\% at 25 demos, +26.0\% at 50 demos, +22.7\% at 200 demos, and +27.0\% at 1000 demos.
Notably, even with 1000 demonstrations, the novel-view success rate increases from 31.0\% to 58.0\%, confirming that EgoDemoGen provides complementary viewpoint coverage that cannot be achieved by simply collecting more demonstrations from the standard egocentric viewpoint.

\begin{table}[t]
  \centering
  \caption{\textbf{Average success rate (\%) vs.\ number of initial standard egocentric viewpoint demonstrations} on four simulation tasks.
  S: standard egocentric viewpoint. N: novel egocentric viewpoint.}
  \label{tab:more_demos}
  \small
  \setlength{\tabcolsep}{2.5pt}
  \resizebox{\columnwidth}{!}{%
  \begin{tabular}{l cc cc cc cc}
      \toprule
      & \multicolumn{2}{c}{\textbf{25}} 
      & \multicolumn{2}{c}{\textbf{50}} 
      & \multicolumn{2}{c}{\textbf{200}} 
      & \multicolumn{2}{c}{\textbf{1000}} \\
      \cmidrule(lr){2-3} \cmidrule(lr){4-5} \cmidrule(lr){6-7} \cmidrule(lr){8-9}
      & S & N & S & N & S & N & S & N \\
      \midrule
      ACT & 38.0 & 12.3 & 66.0 & 20.3 & 75.5 & 32.3 & 75.8 & 31.0 \\
      \rowcolor{gray!15}
      + \textbf{EgoDemoGen} & \textbf{65.0} & \textbf{30.3} & \textbf{71.0} & \textbf{46.3} & \textbf{75.8} & \textbf{55.0} & \textbf{78.0} & \textbf{58.0} \\
      \bottomrule
  \end{tabular}%
  }
\end{table}

\subsection{Generation Efficiency}
\label{sec:supp_efficiency}

\Cref{tab:efficiency} reports the wall-clock time for each stage of the EgoDemoGen pipeline when generating one novel-viewpoint demonstration (100 frames at $320 \times 240$ resolution, measured on a single H20 GPU).

EgoViewTransfer inference dominates the total cost (34.0\,s out of 44.9\,s), while the remaining stages (EgoTrajTransfer, robot motion video rendering, and scene reprojection) are lightweight, taking 2.8\,s, 5.0\,s, and 3.1\,s respectively.
Importantly, the entire pipeline is offline and embarrassingly parallelizable across episodes, making it practical for large-scale data generation.
% For example, generating 25 novel-viewpoint demonstrations from 25 original demonstrations takes approximately 19 minutes on a single GPU, or under 5 minutes with 4 GPUs in parallel.

\begin{table}[t]
  \centering
  \caption{Per-stage wall-clock time for generating one novel-viewpoint demonstration (100 frames, $320 \times 240$, single H20 GPU).}
  \label{tab:efficiency}
  \small
  \setlength{\tabcolsep}{3pt}
  \begin{tabular}{l ccccc}
      \toprule
       & \makecell{EgoTraj-\\Transfer} & \makecell{Robot\\Render} & \makecell{Scene\\Reproj.} & \makecell{EgoView-\\Transfer} & \textbf{Total} \\
      \midrule
      Time (s) & 2.8 & 5.0 & 3.1 & 34.0 & 44.9 \\
      \bottomrule
  \end{tabular}
\end{table}

%% ============================================================
%%  E. Additional Visualizations
%% ============================================================
\section{Additional Visualizations}
\label{sec:supp_vis}

\begin{figure*}[t]
    \centering
    \includegraphics[width=\textwidth]{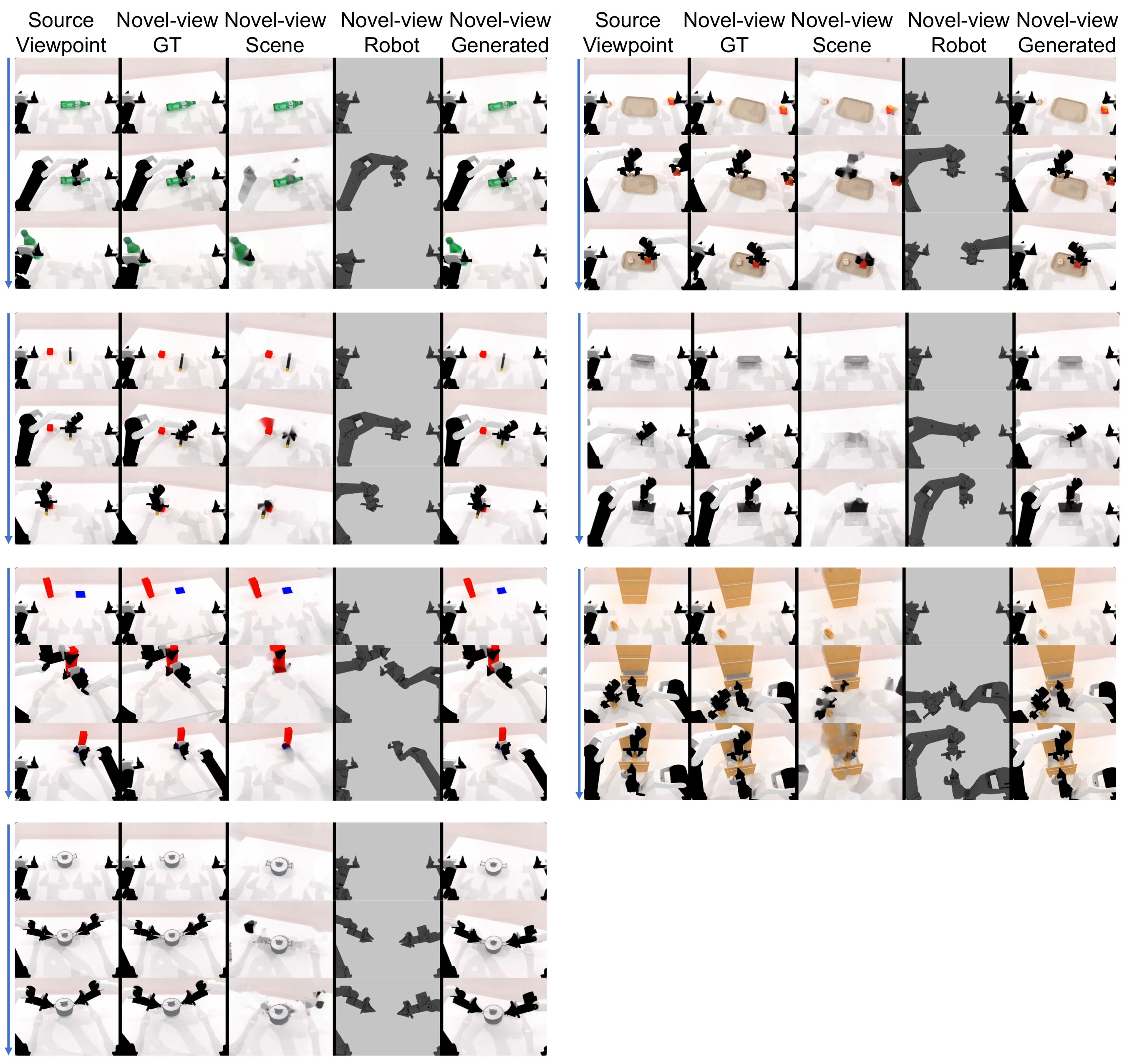}
    \caption{
        \textbf{EgoViewTransfer generation visualization on seven simulation tasks.}
        Each group shows a frame sequence with five rows: Source (standard egocentric viewpoint), GT (ground-truth novel egocentric viewpoint), Scene Video (reprojected novel-viewpoint video with robot masked), Robot Video (rendered robot motion video from transferred trajectory), and Generated (final video synthesized by EgoViewTransfer).
    }
    \label{fig:egoview_vis_sim}
\end{figure*}

\begin{figure*}[t]
    \centering
    \includegraphics[width=\textwidth]{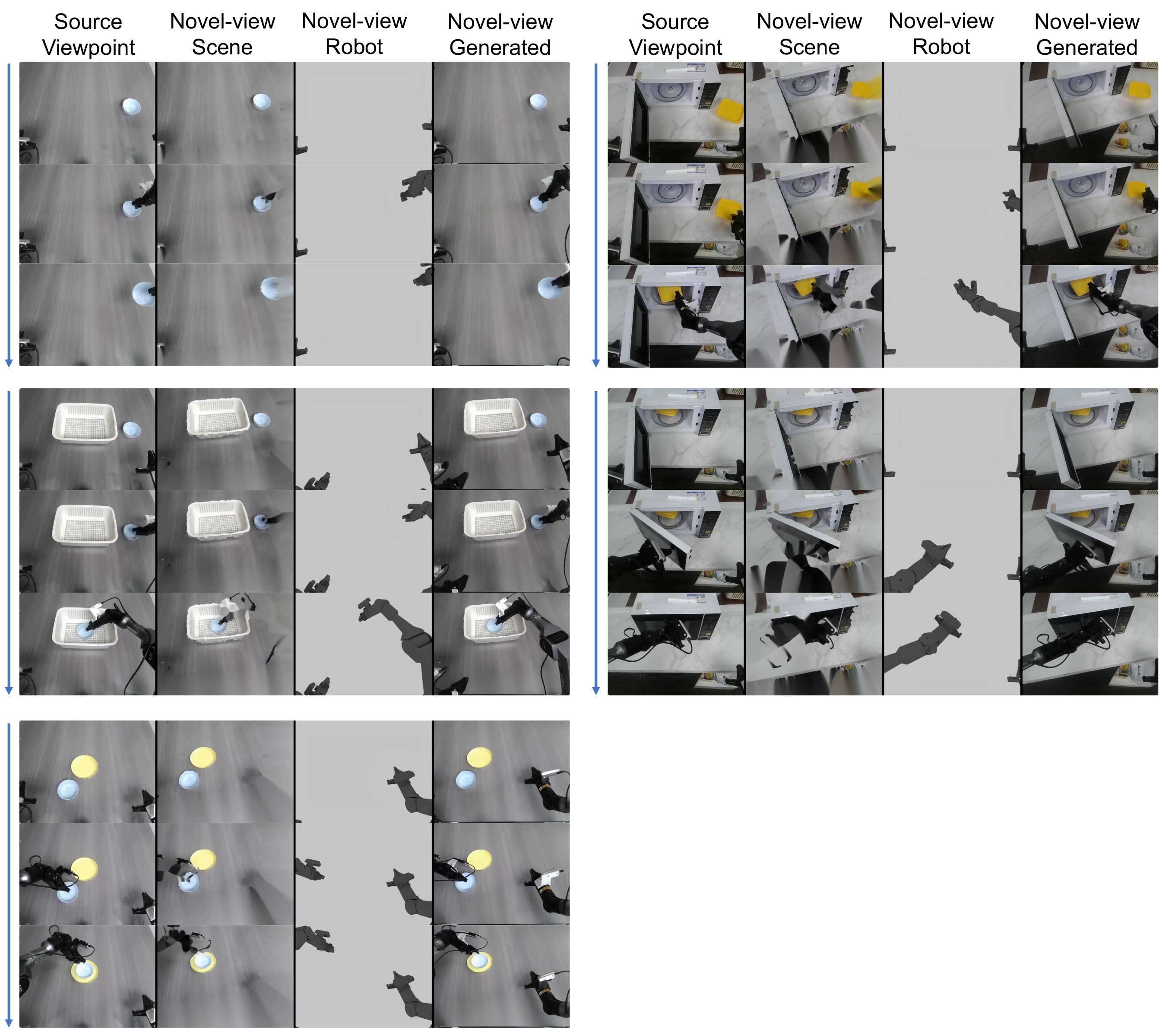}
    \caption{
        \textbf{EgoViewTransfer generation visualization on five real-world tasks.}
        Each group shows a frame sequence with four rows: Source (standard egocentric viewpoint), Scene Video (reprojected novel-viewpoint video with robot masked), Robot Video (rendered robot motion video from transferred trajectory), and Generated (final video synthesized by EgoViewTransfer).
        Ground-truth novel-viewpoint videos are not available in the real-world setting.
    }
    \label{fig:egoview_vis_real}
\end{figure*}

\Cref{fig:egoview_vis_sim} and \Cref{fig:egoview_vis_real} visualize the EgoViewTransfer generation pipeline for all simulation and real-world tasks, respectively.
For each task, we show the intermediate representations and the final generated video as a frame sequence.
In simulation (\Cref{fig:egoview_vis_sim}), each group displays five rows: (1) \textbf{Source}, the original standard egocentric viewpoint video; (2) \textbf{GT}, the ground-truth novel egocentric viewpoint video; (3) \textbf{Scene Video}, the reprojected scene video at the novel viewpoint with the robot masked and inpainted; (4) \textbf{Robot Video}, the rendered robot motion video from the transferred trajectory via EgoTrajTransfer; and (5) \textbf{Generated}, the final novel egocentric viewpoint video synthesized by EgoViewTransfer conditioned on the scene and robot motion videos.
In the real world (\Cref{fig:egoview_vis_real}), ground-truth novel-viewpoint videos are unavailable, so each group shows four rows: Source, Scene Video, Robot Video, and Generated.

The visualizations confirm that EgoViewTransfer faithfully reconstructs the task-relevant scene content while generating realistic robot motion consistent with the transferred trajectory, across diverse manipulation tasks in both simulation and real-world settings.

% {
%     \small
%     \bibliographystyle{ieeenat_fullname}
%     \bibliography{main}
% }
\end{document}